# The Automatic Inference of State Invariants in TIM

**Maria Fox** MARIA.FOX@DUR.AC.UK
**Derek Long** D.P.LONG@DUR.AC.UK
*Department of Computer Science*
*University of Durham, UK*

## Abstract

As planning is applied to larger and richer domains the effort involved in constructing domain descriptions increases and becomes a significant burden on the human application designer. If general planners are to be applied successfully to large and complex domains it is necessary to provide the domain designer with some assistance in building correctly encoded domains. One way of doing this is to provide domain-independent techniques for extracting, from a domain description, knowledge that is implicit in that description and that can assist domain designers in debugging domain descriptions. This knowledge can also be exploited to improve the performance of planners: several researchers have explored the potential of state invariants in speeding up the performance of domain-independent planners. In this paper we describe a process by which state invariants can be extracted from the automatically inferred type structure of a domain. These techniques are being developed for exploitation by STAN, a Graphplan based planner that employs state analysis techniques to enhance its performance.

## 1. Introduction

STAN (Long & Fox, in press) is a domain-independent planner based on the constraint satisfaction technology of Graphplan (Blum & Furst, 1995). Its name is derived from the fact that it performs a variety of pre-processing analyses (STate ANalyses) on the domain description to which it is applied, that assist it in planning efficiently in that domain. STAN took part in the AIPS-98 planning competition, the first international competition in which domain-independent planners were compared in terms of their performance on well-known benchmark domains. Of the four planners that competed in the STRIPS track, three were based on the Graphplan (Blum & Furst, 1995) architecture. The most important difference between STAN and the other Graphplan-based planners was its use of state analysis techniques. Although these techniques were not, at that stage, fully integrated with the planning algorithm STAN gave an impressive performance as can be determined by examination of the competition results. There is a description of the competition, its objectives and the results, at the AIPS-98 planning competition FTP site (see Appendix A).

One of the most important of the analyses performed by STAN is the automatic inference of state invariants. As will be described in this paper, state invariants are inferred from the type structure of the domain that is itself automatically inferred, or enriched, by STAN. The techniques used are completely independent of the planning architecture, so can be isolated in a pre-processing module that we call TIM (Type Inference Module). TIM can be used by any planner, regardless of whether it is based on Graphplan or on any other underlying architecture. TIM has been implemented in C++ and executables and examples of output are available at our web site (see Appendix A) and in Online Appendix 1.

TIM takes a domain description in which no type information need be supplied and infers a rich type structure from the functional relationships between objects in the domain. If type information *is* supplied TIM can exploit it as the foundation of the type structure and will often infer an enriched type structure on this basis. State invariants can be extracted from the way in which the inferred types are partitioned. The consequence is that the domain designer is relieved of a considerable overhead in the description of the domain. Whilst it is easy to hand-code both types and state invariants for simple domains containing few objects and relations, it becomes progressively more difficult to ensure cross-consistency of hand-coded invariants as domains become increasingly complex. Similarly, the exploitable type structure of a domain may be much richer than can easily be provided by hand. We have observed that TIM often infers unexpected type partitions that increase the discrimination of the type structure and provide corresponding benefits to STAN's performance. We therefore see TIM as a domain engineering tool, helping to shift the burden of domain design from the human to the automatic system.

The usefulness of both types and state invariants is well-documented. Types have been provided by hand since it was first observed that they reduce the number of operator instantiations that have to be considered in the traversal of a planner's search space. The elimination of meaningless instantiations is particularly helpful in a system such as Graphplan, in which the structure to be traversed is explicitly constructed prior to search. We believe that the benefits to be obtained from type inference in planning are similar to those obtained in programing language design: type inference is more powerful than type checking and can assist in the identification of semantic errors in the specification of the relational structure of the domain. Indeed, we have found TIM to be a useful domain debugging tool, allowing us to identify flaws in some published benchmark domains. We also used TIM to reveal the underlying structure of the Mystery domain, a disguised transportation problem domain, used in the planning competition. The Mystery domain is described in Appendix C.2.

The use of domain knowledge can significantly improve the performance of planners, as shown by a number of researchers. Gerevini and Schubert (1996a, 1996b) have considered the automatic inference of some state constraints and demonstrated that a significant empirical advantage can be obtained from their use. Kautz and Selman (1998) have hand-coded invariants and provided them as part of the domain description used by Blackbox. They demonstrate the performance advantages obtained and acknowledge the importance of inferring such invariants automatically. McCluskey and Porteous (1997) have also demonstrated the important role that hand-coded state invariants can play in domain compilation for efficient planning. Earlier work by Kelleher and Cohn (1992) and Morris and Feldman (1989) explores the automatic generation of some restricted invariant forms. We discuss these, and other, related approaches in section 5.

In this paper we will describe the type inference process employed by TIM and explain how four different forms of state invariant can be extracted from the inferred type structure. We will argue that TIM is correct since it never infers sentences that are not state invariants. We will then provide experimental results demonstrating the performance advantages that can be obtained by the use of types.

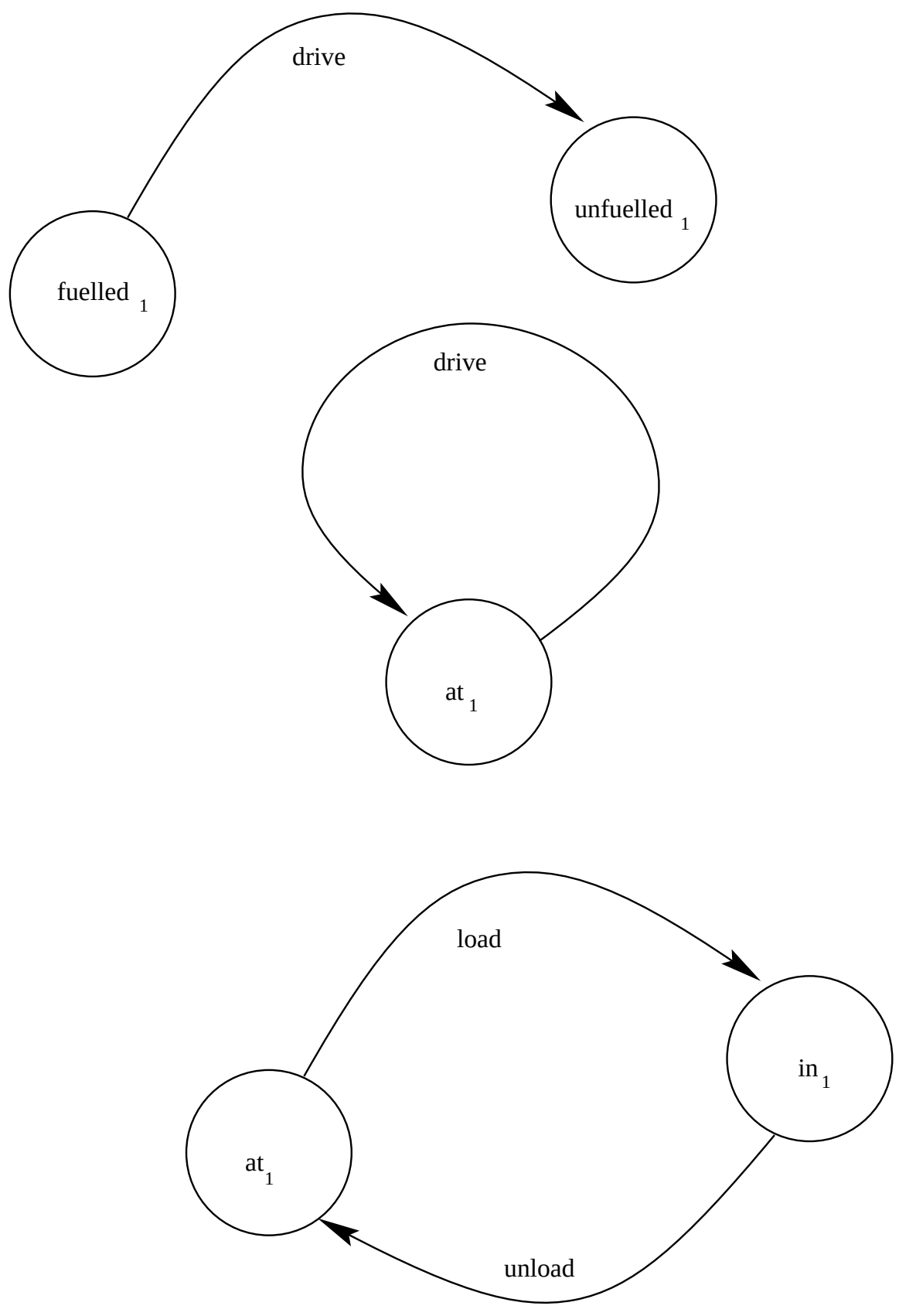

Figure 1: A simple transportation domain seen as a collection of FSMs.

## 2. The Type Inference Module

One way of viewing STRIPS (Fikes & Nilsson, 1971) domains is as a collection of finite-state machines (FSMs) with domain constants traversing the states within them. For example, in a simple transportation domain there are rockets and packages, with rockets being capable of being *at* locations and of moving, by driving, from being *at* one location to being *at* another, and of being *fuelled* or *unfuelled*, and of moving between these two states. *at* can be seen as forming a one-node FSM, and *fuelled* and *unfuelled* as forming a two-node FSM. This view is depicted in Figure 1.

Packages can be *at* locations or *in* rockets, and can move between these states in the resulting two-node FSM. In this example, rockets can be in states that involve more than one FSM, since they can be both *at* and *fuelled*, or *at* and *unfuelled*. STRIPS domains have been seen in this way in earlier work (McCluskey & Porteous, 1997; Grant, 1996), as discussed in Section 5.

### 2.1 Types in TIM

When two objects participate in identical FSMs they are *functionally equivalent* and can be seen to be of the same *type*. The notion of type here is similar to that of *sorts* in the work of McCluskey and Porteous (1997). A primary objective of the TIM module is to automatically identify the equivalence classes that form the primitive types in a domain description and to infer the hierarchical type structure of a domain on the basis of the primitive types. The way this is done is discussed in Section 2.3. The primitive types are functional equivalence classes, and the objects of the domain are partitioned into these classes. Having identified the types of the domain objects TIM infers the types of the parameters of all of the operators. State invariants are inferred as a final stage.

The early parts of this process rely on three key abstract data types, the *property space*, the *attribute space* and the *transition rule*. Formal definitions of these components are provided in Section 2.3, but we provide informal descriptions here to support the following definitions. *Transition rules* represent the state transformations that comprise the FSMs traversed by the objects in the domain. *Property spaces* are FSMs, together with the objects that participate in them, the properties these objects can have and the transition rules by which they can acquire these properties. *Attribute spaces* contain collections of objects that have, or can acquire, the associated attributes. Attributes differ from properties because they can be acquired, or lost, without the associated loss, or acquisition (respectively), of another attribute. Attribute spaces also contain the transition rules that enable the acquisition (or loss) of these attributes. Once the state and attribute spaces have been constructed we assign types to the domain objects according to their membership of the property and attribute spaces. Any two objects that belong in identical property and attribute spaces will be assigned the same type. It is therefore very important to ensure that the property and attribute spaces are *adequately discriminating*, otherwise important type distinctions can be lost. Much of the subtlety of the algorithm described in Section 2.2 is concerned with maintaining adequate discrimination in the construction of these spaces.

We present the following definitions here to support our informal characterisation of the roles of types in STRIPS and in TIM. The definitions are used again in Sections 2.4 and 2.6, which discuss how types are assigned to objects and operator parameters.

**Definition 1** *A* type vector *is a bit vector in which each bit corresponds to membership, or otherwise, of a unique state or attribute space. The number of bits in the vector is always equal to the number of distinct state and attribute spaces.*

**Definition 2** *A* type *is a set of domain objects each associated with the same type vector.*

**Definition 3** *A type vector, $V_1$, in which two distinct bits, $s_i$ and $s_j$, are set corresponds to a* sub-type *of the type associated with a vector, $V_2$, in which only $s_i$ is set (all other settings* *being equal). Then the type associated with* $V_2$ *can be seen to be a* super-type *of the type associated with* $V_1$*.*

**Definition 4** *A* type structure *is a hierarchy of types organised by sub-type relationships between the component types.*

**Definition 5** *A type structure is* adequately discriminating *if objects are only assigned to state (and attribute) spaces that characterize their state transitions (and attributes).*

**Definition 6** *A type structure is* under-discriminating *if it fails to distinguish types that are functionally distinct.*

**Definition 7** *A type structure is* over-discriminating *if functionally identical objects are assigned to different types.*

There are two distinct ways in which types play a role in the specification of a domain. They can restrict the set of possible operator instances to eliminate all those that are meaningless in the domain and hence improve efficiency by reducing the size of the search space, and they can eliminate unsound plans that could be constructed if they were not provided. The following examples clarify the difference between these two roles. The untyped schema:

**drive(X,Y,Z)**

| | |
|---|---|
| Pre: | at(X,Y), fuelled(X), location(Z) |
| Add: | at(X,Z), unfuelled(X) |
| Del: | at(X,Y), fuelled(X) |

permits more instances than the typed schema:

**drive(X,Y,Z)**

| | |
|---|---|
| params: | X:rocket,Y:package,Z:location |
| Pre: | at(X,Y), fuelled(X), location(Z) |
| Add: | at(X,Z), unfuelled(X) |
| Del: | at(X,Y), fuelled(X) |

but all meaningless instances will be eliminated during search because their preconditions will not be satisfiable. On the other hand, the typed schema:

**fly(X,Y,Z)**

| | |
|---|---|
| params: | X:aircraft,Y,Z:location |
| Pre: | at(X,Y) |
| Add: | at(X,Z) |
| Del: | at(X,Y) |

ensures that only aircraft can be flown, whilst the untyped schema:

**fly(X,Y,Z)**

| | |
|---|---|
| Pre: | at(X,Y) |
| Add: | at(X,Z) |
| Del: | at(X,Y) |

allows flying as a means of travel for any object that can be *at* a location, including packages, and other objects, as well as aircraft. TIM is capable of automatically inferring all types playing the *restrictive* role indicated in the typed *drive* operator. However, TIM cannot infer type information that is not implicit in the domain description. Thus, given the untyped *fly* schema, there are no grounds for TIM to infer any type restrictions. TIM will draw attention to unintended under-discrimination by making packages and aircraft indistinguishable at the type level, unless there is distinguishing information provided in other schemas. At the very least TIM will make explicit the fact that packages are amongst those objects that can fly. This assists the domain designer in tracking errors and omissions in a domain description, but unstated *intended* distinctions cannot be enforced by TIM.

### 2.2 An Overview of the TIM Algorithm

Figure 2 gives a broad outline of the TIM algorithm. A more detailed description is given in Appendix B. The role of each component of the algorithm is described, together with a commentary on discussing related issues and justifications, in Sections 2.3, 2.4 and 2.7.

Broadly, TIM begins with an analysis of the domain operators, extracting transition rules that form the foundations of the property and attribute spaces described previously. These rules are used to separate properties into equivalence classes from which the property and attribute spaces are constructed. TIM then analyses the initial state in order to assign the domain objects to their appropriate spaces. This analysis also identifies the initial properties of individual objects and uses them to form *states* of the objects in the property spaces. The initial states in a property space are then extended by the application of the transition rules in that space to form complete sets of states accounting for all of the states that objects in that property space can possibly inhabit. As described in Section 2.4, attribute spaces do not behave like FSMs, as property spaces do, and the extension of these is carried out by a different procedure: one that can add new *objects* to these spaces, rather than new states.

TIM then assigns types to objects using the pattern of membership of the spaces it has constructed. Finally, TIM uses the spaces to determine invariants that govern the behaviour of the domain and the objects in it.

### 2.3 Constructing the Transition Rules

We begin by describing the process by which the transition rules are constructed. The following definitions are required.

**Definition 8** *A* property *is a predicate subscripted by a number between 1 and the arity of that predicate. Every predicate of arity n defines n properties.*

**Definition 9** *A* transition rule *is an expression of the form:*

$$property^* \Rightarrow property^* \rightarrow property^*$$

| |
|---|
| Construct base PRSs (Section 2.3) |
| Split PRSs (Section 2.3) |
| Construct transition rules (Section 2.3) |
| Seed property and attribute spaces (Section 2.3) |
| Assign transition rules (Section 2.4) |
| Analyse initial state (Section 2.4) |
| Extend property spaces (Section 2.4) |
| Extend attribute spaces (Section 2.4) |
| Identify types (Section 2.6) |
| Construct invariants (Section 2.7) |

Figure 2: Outline of the TIM algorithm.

*in which the three components are bags of zero or more properties called* enablers*,* start *and* finish*, respectively.*

The double arrow, $\Rightarrow$, is read *enables* and the single arrow, $\rightarrow$, is read *the transition from*. So:

$$E \Rightarrow S \rightarrow F$$

is read: *E enables the transition from S to F*. The properties in $S$ are given up as a result of the transition. The properties in $F$ are acquired as a result of the transition. The properties in $E$ are not given up.

If *enablers* is empty we write:

$$start \rightarrow finish$$

If *start* is empty we write:

**Transition rule 1**

$$enablers \Rightarrow null \rightarrow finish$$

If *finish* is empty we write:

**Transition rule 2**

$$enablers \Rightarrow start \rightarrow null$$

The bag *null* is the empty bag of properties. Its role is to emphasise that, in transition rule 1, nothing is given up as a result of the transition and, in transition rule 2, nothing is acquired. Rules that have a null start and a null finish are discarded because they describe null transitions.

When the property bags contain more than one element they are separated by commas. The collection:

$$p_k, q_m, \ldots r_n$$

is interpreted to mean that each of the properties in the collection can be satisfied as many times as they appear in the collection. The comma is therefore used to separate the elements of a bag. We use $\oplus$ to denote bag union, $\ominus$ to denote bag difference, $\otimes$ to denote bag intersection and $\sqsubseteq$ to denote bag inclusion.

**Definition 10** *A* Property Relating Structure *(*PRS*) is a triple of bags of properties.*

The first stage of the algorithm constructs a set of transition rules from a set of operator schemas. Each operator schema is analysed with respect to each parameter in turn and, for each parameter, a PRS is built. The first bag of properties is formed from the preconditions of the schema, and the number used to form the property is the argument position of the parameter being considered. For example, if the precondition is $on(X,Y)$, and the parameter being considered is $X$, the property formed is $on_1$. This bag, called *precs*, contains the enablers that will be used in the formation of the transition rules. The second bag, called *deleted_precs*, of properties is formed from all of the preconditions that appear on the delete list of the schema (with respect to this same parameter). The third bag, called *add_elements*, contains the properties that can be formed from the add list of the schema. The PRS contains no *deleted_elements* component – it is assumed that every element on the delete list of a STRIPS operator appears in the precondition list. This is a reasonable restriction given that STRIPS operators do not allow the use of conditional effects. It is further assumed that every pair of atoms on the delete list of a schema will be distinct for all legal instantiations of the schema. This does not constitute a significant restriction since operator schemas can always be easily rephrased whenever this condition is violated.

We now consider the process by which PRSs are constructed. Given the schema:

| **drive(X,Y,Z)** | |
|---|---|
| Pre: | at(X,Y), fuelled(X), location(Z) |
| Add: | at(X,Z), unfuelled(X) |
| Del: | at(X,Y), fuelled(X) |

and considering the parameter $X$, the following PRS will be built:

**PRS 1**

$$\begin{array}{ll} precs: & at_1,\ fuelled_1 \\ deleted_precs: & at_1,\ fuelled_1 \\ add_elements: & at_1,\ unfuelled_1 \end{array}$$

By considering the parameter $Y$ we obtain:

**PRS 2**

$$\begin{array}{ll} precs: & at_2 \\ deleted_precs: & at_2 \\ add_elements: & \end{array}$$

and by considering the parameter $Z$ we obtain:

**PRS 3**

$$\begin{array}{ll} precs: & location_1 \\ deleted_precs: & \\ add_elements: & at_2 \end{array}$$

In constructing these structures we are identifying the state transformations through which the objects, instantiating the operator parameters, progress. Note that objects that instantiate $X$ go from being *fuelled* and *at* somewhere to being *unfuelled* and *at* somewhere; objects that instantiate $Y$ lose the property of having anything *at* them and gain nothing as a result of application of this operator, and objects that instantiate $Z$ continue being *locations* and gain the property of having something *at* them. We now convert these structures into transition rules in order to correctly capture these state transformations.

Our standard formula for the construction of rules from PRSs is:

$$precs \ominus deleted_precs \Rightarrow deleted_precs \rightarrow add_elements$$

Thus, using the PRS 1 above, we could build the rule:

$$at_1,\, fuelled_1 \rightarrow at_1,\, unfuelled_1$$

A potential problem with this rule is that it causes $at_1$ and $fuelled_1$ to be linked in state transformations, so that $at_1$ and $fuelled_1$ become associated with the same property space and, as a consequence, objects that can be *at* places, but that cannot be *fuelled*, may be indistinguishable from objects that require fuelling before they can be moved. In fact, we wish the transition rules to express the fact that being *fuelled* enables things to go from being *at* one place to being *at* another place, whilst not excluding the possibility that there may be other enablers of this transition.

We therefore begin a second phase of PRS construction by identifying, for special treatment, PRSs in which a property appears in both the *deleted_precs* and the *add_elements*. This is a property that is *exchanged* on application of the operator. That is, the relation continues to hold between the identified argument and some other object or objects (not necessarily the same object or objects as before the application of the operator). For example, in PRS 1, the vehicle is *at* a new location after application of the operator, and no longer *at* the old location. We observe that the vehicle must be *fuelled* to make this transition. To separate the transition from this condition we *split* the PRS. Splitting identifies the exchanged properties in a PRS and creates one new PRS for each exchange and one for the unexchanged properties. Therefore, splitting a PRS always results in at most $k+1$ (and at least $k$) new PRSs, where $k$ is the number of exchanges that the PRS represents. By splitting PRS 1 we construct two new PRSs: one characterizing the exchange of the *at* property, and one characterising the *fuelled* to *unfuelled* transition.

The first of the new PRSs is:

**PRS 4**

$$\begin{array}{ll} precs: & at_1,\, fuelled_1 \\ deleted_precs: & at_1 \\ add_elements: & at_1 \end{array}$$

from which the rule

$$fuelled_1 \Rightarrow at_1 \rightarrow at_1$$

is constructed. It should be noted that the property of being *fuelled* is no longer seen as part of the state transformation but only as an enabler, which is why it does not appear in the *deleted_precs* bag in the resulting PRS.

The second new PRS captures the fact that $at_1$ can be seen as an enabler for the transition from $fuelled_1$ to $unfuelled_1$:

**PRS 5**

$$\begin{array}{ll} precs: & at_1,\, fuelled_1 \\ deleted_precs: & fuelled_1 \\ add_elements: & unfuelled_1 \end{array}$$

In this PRS there are no further splits required since no other properties are exchanged in it. A more general example is as follows:

**PRS 6**

$$\begin{array}{ll} precs: & p_1,\, p_2 \cdots p_n \\ deleted_precs: & p_1 \cdots p_i\, p_{i+k} \cdots p_m \\ add_elements: & p_1 \cdots p_i\, q_1 \cdots q_k \end{array}$$

from which $i$ PRSs would be constructed to deal with each of the $i$ exchanged pairs and a final PRS, PRS 7, would be constructed to describe the remainder of the transition making $i+1$ PRSs in total.

**PRS 7**

$$\begin{array}{ll} precs: & p_1,\, p_2 \cdots p_n \\ deleted_precs: & p_{i+k} \cdots p_m \\ add_elements: & q_1 \cdots q_k \end{array}$$

There is no need to consider additional pairings of add and delete-list elements, since these would not correspond to exchanges of properties. The splitting process is justified in Section 3.1. The standard rule construction formula can be applied to PRS 5, yielding the rule

$$at_1 \Rightarrow fuelled_1 \rightarrow unfuelled_1$$

It should be observed that, even if the *add_elements* bag contains multiple properties, a single rule will always be built when the standard construction formula is applied.

On considering the remaining PRSs, 2 and 3, it can be observed that they each contain an empty field: in 2 the *add_elements* field is empty and in 3 the *deleted_precs* field is empty. When a PRS has an empty field special treatment is required. From PRS 2 we build the rule

$$at_2 \rightarrow null$$

to represent the fact that the object that instantiates $Y$ gives up the property of having something *at* it, and gains nothing in return. From 3 we build the rule

$$location_1 \Rightarrow null \rightarrow at_2$$

to represent the fact that the object that instantiates $Z$ gains the property of having something *at* it by virtue of being a *location*, and gives up nothing in return. These rules have a somewhat different status from the ones that characterize the exchange of properties. In these cases properties are being lost or gained, without exchange, so can be seen as *resources* that can be accumulated or spent by domain objects rather than as states through which the domain objects pass. For example, a location can acquire the property of having something *at* it, without relinquishing anything in return, whereas an object that requires fuel can only become *fuelled* by relinquishing the property of being *unfuelled*, and vice versa. Increasing and decreasing resources are identified as *attributes* and are distinguished from states. This distinction will later prove to be very important, since the generation of true state invariants depends upon it being made correctly. Properties that can increase and decrease without exchange are not invariant, and false assertions would be proposed as invariants if they were treated in the same way as state-valued properties.

A rule of the form constructed from PRS 3 must be constructed separately for every property in the *add_elements* bag because these properties must be individually characterized as increasing resources. Rules constructed using *null* are distinguished as *attribute transition rules*. If the *null* is on the left side of the $\rightarrow$ the rule is an *increasing attribute transition rule*. If the *null* is on the right hand side then the rule is a *decreasing attribute transition rule*.

A final case to consider during rule construction is the case in which a PRS has an empty *precs* field. This happens if the parameter, with respect to which the PRS was constructed, did not appear in any of the preconditions of the operator schema. In this case a set of rules is constructed, one for each property, $a$, in the *add_elements* bag, of the form

$$null \rightarrow a$$

reflecting the fact that $a$ is an increasing resource (the *deleted_precs* field will necessarily also be empty in this case).

**Definition 11** *A* state *is a bag of properties.*

When it is necessary to distinguish a bag from a set, square brackets will be used to denote the bag.

**Definition 12** *A* property space *is a tuple of four components: a set of properties, a set of transition rules, a set of states and a set of domain constants.*

**Definition 13** *An* attribute space *is a tuple of three components: a set of properties, a set of transition rules and a set of domain constants.*

It is helpful to observe here that the state and attribute spaces represent disjoint collections of properties, and that these disjoint collections are formed from the transition rules by putting the *start* and *finish* properties of each rule into the same collection. For example, given two rules:

$$E_1 \Rightarrow [p_1, p_2, p_3] \rightarrow [q_1, q_2]$$

and

$$E_2 \Rightarrow [r_1, r_2] \rightarrow [s_1]$$

the collections $[p_1, p_2, p_3, q_1, q_2]$ and $[r_1, r_2, s_1]$ would be formed. If a property appears in the *start* or *finish* of both rules then a single collection will be formed from the two rules.

The last stage in the rule construction phase is to identify the basis for the construction of property and attribute spaces. This is done by *uniting* the left and right hand sides of the rules. Uniting forms collections of properties that each seed a unique property or attribute space. It is not yet possible to decide which of the seeds will form attribute spaces, so treatment of both kinds of space is identical at this stage. The enablers of the rules are ignored during this process. We do not wish to make enablers automatically fall into the same property spaces as the states in the transformations they enable. This could result in incorrect assignment of properties to property and attribute spaces since enablers only facilitate, and do not participate in, state transformations. The output of this phase is the collection of rules, with some properties marked as attributes, and the property space seeds formed from the uniting process. All properties that remain unassigned at this stage are used to seed separate attribute spaces, one for each such property.

The role played by the second phase of PRS construction is to postpone commitment to the uniting of collections of properties so that the possibility of objects, which can have these properties, being associated with different property spaces is left open for as long as possible. It may be that consideration of other schemas provides enough information for this possibility to be eliminated, as in the following abstract example, but we support as much type discrimination as possible in the earlier phases of analysis. We consider this simple example to illustrate the problem.

#### 2.3.1 Postponing Property Space Amalgamation

Given a domain description containing the following operator schema:

| **op1(X,Y,Z)** | |
|---|---|
| Pre: | p(X,Y), q(X,Y) |
| Add: | p(X,Z), q(X,Z) |
| Del: | p(X,Y), q(X,Y) |

the PRS:

$$\begin{array}{ll} precs: & p_1,\ q_1 \\ deleted_precs: & p_1,\ q_1 \\ add_elements: & p_1,\ q_1 \end{array}$$

will be constructed, during the first phase, for $X$. The properties $p_1$ and $q_1$ are bound together in this PRS, and the resulting rule would be:

$$p_1, q_1 \rightarrow p_1, q_1$$

which forces objects that can have property $p_1$ to occupy the same property space as objects that can have property $q_1$. Since this PRS models the exchange of $p_1$ we will split it, and replace it with two new PRSs:

$$\begin{array}{ll} precs: & p_1,\ q_1 \\ deleted_precs: & p_1 \\ add_elements: & p_1 \end{array}$$

| | |
|---|---|
| $precs:$ | $p_1, q_1$ |
| $deleted_precs:$ | $q_1$ |
| $add_elements:$ | $q_1$ |

We do not consider other pairings of $p_1$ and $q_1$, since these will be found in the PRSs of other operator schemas if the domain allows them. The two PRSs generated lead to the generation of the rules:

$$q_1 \Rightarrow p_1 \rightarrow p_1$$

and

$$p_1 \Rightarrow q_1 \rightarrow q_1$$

The two rules indicate that $p_1$ and $q_1$ should be used to form different property spaces since they could, in principle, be independent of one another. Then objects assigned to these two spaces can turn out to be of distinct types. However, if we add the following two schemas:

| **op2(X,Y)** | | **op3(X,Y,Z)** | |
|---|---|---|---|
| Pre: | q(X,Y) | Pre: | p(X,Y) |
| Add: | p(X,Y) | Add: | q(X,Y) |
| Del: | q(X,Y) | Del: | p(X,Y) |

we generate, for $X$, the PRSs:

| | |
|---|---|
| $precs:$ | $q_1$ |
| $deleted_precs:$ | $q_1$ |
| $add_elements:$ | $p_1$ |

and

| | |
|---|---|
| $precs:$ | $p_1$ |
| $deleted_precs:$ | $p_1$ |
| $add_elements:$ | $q_1$ |

and the rules:

$$q_1 \rightarrow p_1$$

and

$$p_1 \rightarrow q_1$$

indicating that $p_1$ and $q_1$ should be united in the same set and hence form a single property space, and that objects that can have these properties are really of the same type. The uniting overrides the potential for separate property spaces to be formed but, in the absence of these two schemas, there would have been insufficient information available to determine the nature of the relationship between the two properties.

### 2.4 Constructing the Property Spaces and Synthesising the Types

The objective of this stage is to construct the type structure of the domain by identifying domain objects with distinct property spaces. Objects can appear in more than one property space, giving us a basis for deriving a hierarchical type structure.

The first part of the process involves completing the seeded property spaces. The first task is to associate transition rules with the appropriate property space seeds. This can be easily done by picking an arbitrary property of the *start* or *finish* component of each rule and identifying the property space seed to which that property belongs. There can never be ambiguity because every property belongs to only one seed and uniting ensures that all of the properties referred to in a rule belong to the same seed. At this point the distinction between states and attributes becomes important. Any property space seed that has an attribute transition rule associated with it becomes an *attribute space* and is dealt with differently from property spaces in certain respects explained below.

The next step is to identify the domain objects associated with each property space and attribute space.

For each object referred to in the initial state we construct a type vector in which a bit is set if the corresponding space is inhabited by the object. An object can inhabit more than one space. Habitation is checked for by identifying all of the properties that hold, in the initial state, of the object being considered and allocating them as *states*, rather than as *properties*, to the appropriate state and attribute spaces. When every domain object has been considered a unique type identifier is associated with each of the different bit patterns.

The next task is to populate the property spaces with states. The following definitions are required to support the explanation of this process.

**Definition 14** *A* world-state *is a collection of propositions characterising the configuration of the objects in a given planning domain description.*

**Definition 15** *Given a world-state, $W$, a property space, $P = (Ps, TRs, Ss, Os)$, or an attribute space, $P = (Ps, TRs, Os)$, and an object $o \in Os$, the* $P$-projection of $St$ for $o$ *is the bag of properties, possessed by $o$ in $W$, each of which belongs to $Ps$.*

The collection of properties of an object, $o$, in the initial state can be divided into a set of bags of properties, each bag corresponding to the $P$-projection of the initial state for $o$, for some property or attribute space $P$. Each bag is added to the state set of the corresponding property space, or discarded if the corresponding space is an attribute space. We now need to extend the spaces by, for each property space, adding states that can be inferred as reachable by objects within that space along transitions within that space. This is done for every state in the space, including states that are newly added during this process, until no further new states are reachable. The ordering of the properties within states is irrelevant, so two states are considered equal if they contain the same properties, regardless of ordering (they are considered *order-equivalent*). Since, when we come to use this information in parts of the process of invariant generation, we will not require knowledge of any inclusion relations between pairs of states, it is convenient to mark these at this stage. The addition of reachable states is important for the inference of state invariants, and their use will be discussed in Section 2.7. The attribute spaces receive different treatment at this point. The important difference to observe is that, since property spaces characterize the exchange of properties, objects in a property space must start off in the initial state as members of that property space. However, since attributes can be acquired without exchange, it is possible for objects that do not have particular attributes in the initial state to acquire those attributes later. This is only possible if the attribute space has an increasing attribute transition rule associated with it. We now, therefore, consider each attribute space to see whether further objects can be added by application of any corresponding increasing rule.

An object can be added to an attribute space if it *potentiates* all of the enablers of an increasing rule in that attribute space. An object potentiates an enabling property if it is a member of the state or attribute space to which that property belongs. Membership of all of these spaces indicates that the object could enter a state in which it satisfies all of the enabling properties, which would justify an application of the increasing rule. Any enabling property that is not associated with a state or attribute space is a static condition, so the initial state can be checked to confirm that the property is true of the object being considered.

A complication arises if any enabling property was itself used to seed an attribute space (in which case it is itself an attribute), because it is then necessary to identify all of the objects in *its* attribute space and consider them for addition to the current attribute space. Of course this could, in principle, initiate a loop in the process but we avoid this by marking attribute spaces as they are considered and ensuring, by iterating until convergence, that all of the attribute spaces in the loop are completely assigned. The correctness of this part of the procedure is discussed in Section 3.

When this is done the state and attribute spaces are complete and the types of the domain objects can be extracted. The completeness of this construction phase is discussed in Section 3.1.

### 2.5 A Worked Example

A fully worked example of all stages of the process will help to clarify what is involved. Consider a simplified version of the Rocket domain in which there are two operator schemas:

**drive(X,Y,Z)**

| | |
|---|---|
| Pre: | at(X,Y), fuelled(X), location(Z) |
| Add: | at(X,Z), unfuelled(X) |
| Del: | at(X,Y), fuelled(X) |

**load(X,Y,Z)**

| | |
|---|---|
| Pre: | at(X,Y), at(Z,Y) |
| Add: | in(X,Z) |
| Del: | at(X,Y) |

and an initial state containing four constants: *rocket, package, London* and *Paris*, and the relations: *at(rocket,Paris), fuelled(rocket)* and *at(package,London)*. It can be observed that this simplified Rocket domain has the rather odd feature that the *load* schema is not restricted to loading *packages* into *rockets*. This oddity will be highlighted by the analysis that is constructed, showing how the analysis performed by TIM can help in understanding (and debugging) the behaviour of the domain. From the *drive* operator schema the following PRSs are constructed for variables $X$, $Y$ and $Z$ respectively:

precs: $at_1$, $fuelled_1$
deleted_precs: $at_1$, $fuelled_1$
add_elements: $at_1$, $unfuelled_1$

precs: $at_2$
deleted_precs: $at_2$
add_elements:

precs: $location_1$
deleted_precs:
add_elements: $at_2$

From the *load* operator schema the following PRSs are constructed for variables $X$, $Y$ and $Z$ respectively:

precs: $at_1$
deleted_precs: $at_1$
add_elements: $in_1$

precs: $at_2$, $at_2$
deleted_precs: $at_2$
add_elements:

precs: $at_1$
deleted_precs:
add_elements: $in_2$

and the following rules are built. The first PRS generates the first two rules and subsequent PRSs each generate one rule.

$$fuelled_1 \Rightarrow at_1 \rightarrow at_1$$
$$at_1 \Rightarrow fuelled_1 \rightarrow unfuelled_1$$
$$at_2 \rightarrow null$$
$$location_1 \Rightarrow null \rightarrow at_2$$
$$at_1 \rightarrow in_1$$
$$at_2 \Rightarrow at_2 \rightarrow null$$
$$at_1 \Rightarrow null \rightarrow in_2$$

We now construct the following united sets of properties:

$$\{at_1, in_1\}$$
$$\{fuelled_1, unfuelled_1\}$$
$$\{at_2\}$$
$$\{in_2\}$$

These are used to seed property spaces. We first associate the rules with these property space seeds, resulting in the following assignment:

| | |
|---|---|
| $\{at_1, in_1\}$ | $at_1 \rightarrow in_1$, $fuelled_1 \Rightarrow at_1 \rightarrow at_1$ |
| $\{fuelled_1, unfuelled_1\}$ | $at_1 \Rightarrow fuelled_1 \rightarrow unfuelled_1$ |
| $\{at_2\}$ | $location_1 \Rightarrow null \rightarrow at_2$, $at_2 \Rightarrow at_2 \rightarrow null$, $at_2 \rightarrow null$ |
| $\{in_2\}$ | $at_1 \Rightarrow null \rightarrow in_2$ |

The last two spaces have been converted into attribute spaces by their association with attribute transition rules. The resulting spaces can now be supplemented with domain constants and their legal states. We first identify the subset of the legal states of the domain objects that are identifiable from the initial state. We do not use the goal state to provide further information about the properties of objects. The goal state might be unachievable because objects cannot obtain the required properties. This would invalidate TIM's analysis of the domain. In the initial state the *rocket* has properties $at_1$ and $fuelled_1$, the *package* has property $at_1$, *London* has property $at_2$ and *Paris* has property $at_2$. Using this information we associate domain constants with the developing state and attribute spaces to obtain:

| | | |
|---|---|---|
| $\{at_1, in_1\}$ | $at_1 \rightarrow in_1$, $fuelled_1 \Rightarrow at_1 \rightarrow at_1$ | $\{rocket, package\}$ |
| $\{fuelled_1, unfuelled_1\}$ | $at_1 \Rightarrow fuelled_1 \rightarrow unfuelled_1$ | $\{rocket\}$ |
| $\{at_2\}$ | $location_1 \Rightarrow null \rightarrow at_2$, $at_2 \Rightarrow at_2 \rightarrow null$, $at_2 \rightarrow null$ | $\{London, Paris\}$ |
| $\{in_2\}$ | $at_1 \Rightarrow null \rightarrow in_2$ | |

The next step is to add the legal states of these objects, which are identifiable so far, to the property spaces. This results in the following structures, the first two of which can be extended by inference (as will be explained) into completed property spaces. The last two will be extended into completed attribute spaces by the addition of objects that can potentially acquire the associated attributes (also described below).

| | | |
|---|---|---|
| $\{at_1, in_1\}$ | $at_1 \rightarrow in_1$, $fuelled_1 \Rightarrow at_1 \rightarrow at_1$ | $\{rocket, package\}$ $[at_1]$ |
| $\{fuelled_1, unfuelled_1\}$ | $at_1 \Rightarrow fuelled_1 \rightarrow unfuelled_1$ | $\{rocket\}$ $[fuelled_1]$ |
| $\{at_2\}$ | $location_1 \Rightarrow null \rightarrow at_2$, $at_2 \Rightarrow at_2 \rightarrow null$, $at_2 \rightarrow null$ | $\{London, Paris\}$ |
| $\{in_2\}$ | $at_1 \Rightarrow null \rightarrow in_2$ | |

The last stage in the construction of the two property spaces is to add any states that can be inferred as reachable, via transition rules, by objects in the property spaces. For example, packages can go from being $at_1$ to being $in_1$, by application of the rule $at_1 \rightarrow in_1$, and since that rule is available in the property space to which package belongs, and $at_1$ is one of the legal states in that property space, we add $in_1$ as a further legal state. In general, we construct the extension by, for each state in the space, identifying applicable rules and, for each rule, creating a new state by removing the properties in the start of the rule and adding the properties in the finish of the rule. This is done until all further states are order-equivalent to those already generated. The enablers of the rules are ignored, with the consequence that some of the new states generated might be unreachable. When this process is completed in the current example the finished property spaces are as follows:

**Property space 1**

$$\{at_1, in_1\} \quad at_1 \rightarrow in_1,\ fuelled_1 \Rightarrow at_1 \rightarrow at_1 \quad \{rocket, package\}$$
$$[at_1], [in_1]$$

**Property space 2**

$$\{fuelled_1, unfuelled_1\} \quad at_1 \Rightarrow fuelled_1 \rightarrow unfuelled_1 \quad \{rocket\}$$
$$[fuelled_1], [unfuelled_1]$$

We now consider each attribute space in turn and add domain objects (not already members) that potentiate their increasing rules. No new domain objects can be added to the first attribute space since only *London* and *Paris* can potentiate the increasing rule, and they are already present. However, when the second attribute space is considered it can be observed that *rocket* and *package* both potentiate the increasing rule and are therefore both added as new members. The resulting attribute spaces are:

$$\{at_2\} \quad location_1 \Rightarrow null \rightarrow at_2,\ at_2 \Rightarrow at_2 \rightarrow null,\ at_2 \rightarrow null \quad \{London, Paris\}$$
$$\{in_2\} \quad at_1 \Rightarrow null \rightarrow in_2 \quad \{rocket, package\}$$

The oddity of the *load* operator is revealed at this stage, since both *package* and *rocket* have been assigned as members of the $in_2$ attribute space (meaning that they both can have the attribute of having things *in* them).

The number of distinct bit patterns that are constructed, indicating object membership of the state and attribute spaces, determines the number of distinct types that exist in the domain. Hence, in this simplified encoding of the Rocket domain, there are three distinct types. The *rocket* has type [1101], the *package* has type [1001] and *Paris* and *London* both have type [0010]. These types are given abstract identifiers, $T_0$, $T_1$ and $T_2$, but might be more meaningfully interpreted as the types of: *movable object requiring fuel, movable object* and *location* respectively. As expected, *London* and *Paris* are of type *location*, whilst the *package* is of type *movable object* and the *rocket* is of type *movable object requiring fuel*, which is a sub-type of *movable object.*

The distinction we have made between state and attribute spaces is further exploited in the process of inferring state invariants, discussed in Section 2.7.

### 2.6 The Assignment of Types to Operator Parameters

Types are assigned to the parameters of the operators in the following way. Given an operator schema and a collection of property spaces and attribute spaces we allocate a type vector to each of the variables in the schema. The membership in the state and attribute spaces of each of the properties of a given variable is recorded by setting the appropriate bits in the vector for that variable. Only the properties that appear in the preconditions of the schema are considered, because any object that can satisfy the preconditions of an operator can have the properties represented by the postconditions and is therefore of the right type for instantiation of the operator. When a type is associated with the vector the union of all of its sub-types is taken. This union is then the type assigned to the variable. Any domain object, the type of which is a sub-type of the type associated with the variable, can then be used to instantiate that variable. To see how this process works, consider the variable $X$ in the *drive* schema above. The precondition properties of $X$ are: $at_1$, $fuelled_1$. These are members of the two property spaces 1 and 2. Therefore, the type vector associated with $X$ is [1100]. It can be observed that the type vector associated with the *rocket* is [1101], so that the type of *rocket* is a sub-type of the type of $X$. This is the only sub-type, so the union of sub-types contains only $T_0$, the type of *rocket*. This means that $X$ can be instantiated by *rocket*, but not by any other domain constant, since no other domain constant has a type in the appropriate sub-type relation. To type the operator parameters we introduce new type variables, $T_k..T_n$ for unused values between $k$ and $n$, where $k$ is the number of existing types and $n$ is $k$ plus the number of variables in the schema being considered. The type vector for variable $Y$ will be [0010] and $Z$ will have no type vector because *location* is a static relation and $Z$ does not appear as an argument to any other predicate in the preconditions. $Z$ therefore acquires the same type as *London* and *Paris*, the only two objects for which *location* is true in the initial state. $T_4$ is a super-type of $T_2$. After taking the unions of the sub-types we can now specify the *drive* schema in the following way:

| **drive(X,Y,Z)** | |
|---|---|
| Params: | X:$T_0$ Y:$T_2$ Z:$T_2$ |
| Pre: | at(X,Y), fuelled(X), location(Z) |
| Add: | at(X,Z), unfuelled(X) |
| Del: | at(X,Y), fuelled(X) |

STAN exploits the sub-typing relations that have been inferred when constructing instances of the *drive* operator. Any variable that appears in a schema but does not appear in its preconditions can be instantiated by objects of any type. This is because the domain description contains no basis for inferring type restrictions in this case. No variable can appear on the delete list without appearing on the precondition list, since we assume that all delete list elements appear as preconditions. So such a variable would have to occur on the add list. This would mean that, regardless of the properties holding of the object used to instantiate that variable, in the initial state, it can acquire that add list property freely. Since this acquisition would occur irrespective of the type of the object, such variables are essentially polymorphic.

### 2.7 The Inference of State Invariants

The final phase of the computation of TIM is the inference of the state invariants from the property spaces. The attribute spaces are not used for the inference of invariants: incorrect invariants would be proposed by TIM if attribute spaces were inadvertantly used. This explains the importance of identifying the attribute spaces in the earlier stages of the algorithm.

The current version of TIM is capable of inferring four kinds of invariant, three of which are inferred from the property spaces (identity invariants, state membership invariants and invariants characterizing uniqueness of state membership) and one of which is inferred from the operator schemas and initial state directly (fixed resource invariants). In the simplified Rocket domain, considered above, an example of an identity invariant is:

$$\forall x : T_k.\forall y.\forall z.(at(x,y) \wedge at(x,z) \rightarrow x = z)$$

A state membership invariant is:

$$\forall x : T_k.(\exists y : T_n.at(x,y) \vee \exists y : T_m.in(x,y))$$

A uniqueness invariant is:

$$\forall x : T_k.\neg(\exists y : T_n.at(x,y) \wedge \exists y : T_m.in(x,y))$$

To infer the identity invariants each property space is considered in turn, with respect to their properties and states. If a property, for example $P_k$ with $P$ of arity $n > 1$, occurs at most once in any state an invariant of the following form, in which $\bar{y}$ and $\bar{z}$ are vectors containing $n-1$ values, can be constructed:

$$\forall x.\forall \bar{y}.\forall \bar{z}.(P(\overline{y_{1..k-1}}, x, \overline{y_{k..n-1}}) \wedge P(\overline{z_{1..k-1}}, x, \overline{z_{k..n-1}}) \rightarrow \bar{y} = \bar{z})$$

The form of this invariant can be generalised to deal with the case where there are at most $m > 1$ occurrences of $P_k$ in any state in the space. In this case we build the following expression, in which we have assumed that $k = 1$, for simplicity.

$$\forall x.\forall \bar{y}_1 ... \bar{y}_m.(P(x,\bar{y}_1) \wedge ... \wedge P(x,\bar{y}_m) \rightarrow (\bar{y}_1 = \bar{y}_2 \vee \bar{y}_1 = \bar{y}_3 \vee ... \vee \bar{y}_{m-1} = \bar{y}_m))$$

The state membership invariants are of the form:

$$\forall x.(Disjunct_1 \vee .. \vee Disjunct_n)$$

where each disjunct is constructed from a single state. Thus, if a property space contains $k$ states there will be at most $k$ disjuncts in the invariant constructed for that property space. Only one state membership invariant is constructed for each property space.

Given the collection of states in a property space we first identify those that are supersets of other states in the collection. All supersets are discarded, since the invariants that would be built from them would be logically equivalent to those built from their subset states. Each remaining state is used to build a single disjunct. If the state being considered contains a single property, $P_k$ with $P$ of arity $n$, then the expression

$$\exists \bar{y}.P(\overline{y_{1..k-1}}, x, \overline{y_{k..n-1}})$$

is constructed. Of course, if $n = 1$ then there is no existential quantifier and the disjunct is just $P(x)$. If the state contains more than one property, say $m$ of them denoted $P^1..P^m$, then we build (again, assuming that $k = 1$ for simplicity):

$$\exists \bar{y}_1 ... \bar{y}_m.(P^1(x,\bar{y}_1) \wedge P^2(x,\bar{y}_2) \wedge ... \wedge P^m(x,\bar{y}_m))$$

The uniqueness invariants are constructed in a similar way. For each property space we begin by analysing the superset states to identify non-exclusive pairs of subset states. For example, given the subset states $\{at_1\}$ and $\{in_1\}$ and the superset state $\{at_1, in_1\}$, it can be observed that the two subset states are not mutually exclusive since $at_1$ and $in_1$ can be simultaneously held. Having done this analysis and identified all mutually exclusive pairs of states we mark the subset states as unusable for generation of invariants. The remaining states are considered in all possible pairings. For every pair of states, $P, Q$, we generate an invariant of the following form assuming, for simplicity, that $x$ is in the first position in $P^1..P^n$ and $Q^1..Q^m$. The form of the invariant is easily generalised, as before.

$$\forall x.\neg(\exists \bar{y}_1...\bar{y}_n.(P^1(x,\bar{y}_1) \wedge P^2(x,\bar{y}_2) \wedge ... \wedge P^n(x,\bar{y}_n)) \\ \wedge(\exists \bar{y}_1...\bar{y}_m.(Q^1(x,\bar{y}_1) \wedge Q^2(x,\bar{y}_2) \wedge ... \wedge Q^m(x,\bar{y}_m))))$$

The fourth kind of invariant can be inferred from the structure of the operator schemas without reference to the property spaces or domain type structure. We call these invariants *fixed resource invariants* since they capture the physical limitations of the domain. Fixed resource invariants cannot be inferred from the state and attribute spaces because they describe properties of the domain rather than of objects within it. The following schema from the Gripper domain provides an example of why fixed resource invariants are distinguished from the other three kinds:

**move(X,Y)**

| | | |
|---|---|---|
| Pre: | at_robot(X), | room(Y) |
| Add: | at_robot(Y) | |
| Del: | at_robot(X) | |

The PRSs that would be built from this operator are:

$$\begin{array}{ll} precs: & at_robot_1 \\ deleted_precs: & at_robot_1 \\ add_elements: & \end{array}$$

$$\begin{array}{ll} precs: & room_1 \\ deleted_precs: & \\ add_elements: & at_robot_1 \end{array}$$

and the rules constructed from these are:

$$at_robot_1 \rightarrow null$$

and

$$room_1 \Rightarrow null \rightarrow at_robot_1$$

It can be observed that both of these rules are attribute transition rules and that $at_robot_1$ is attribute rather than state-valued. This means that no invariants of the first three kinds discussed would be constructed.

The reason for the lack of invariants of the first three forms is that the encoding of the *robot* is embedded in a predicate, so the *robot* cannot participate directly in state transitions. An obvious invariant of the *robot*, which would naturally be true of this domain, is that the *robot* is always in exactly one room but this cannot be inferred using the techniques so far described. In fact, this is an axiom about the *world*, or domain, rather than specific objects within it, and has to be obtained from information other than the state transformations of the objects.

It can be seen from the operator schemas for the Gripper domain that $at_robot_1$ is balanced. That is, it is always deleted whenever it is added and added whenever it is deleted. This means that the number of occurrences of *at_robot* in the initial state determines the number of occurrences that are possible in any subsequent state. This leads to the construction, for this domain, of the invariant

$$|\{x : at_robot(x)\}| = 1$$

since there is only one *at_robot* relation in the initial state. The form of fixed resource invariants is always equational. Such an invariant states that the size of the set of combinations of objects satisfying a certain predicate is equal (or, in some cases, less than or equal) to a certain positive integer. Because this integer can be very large it is more convenient to write an equation than it would be to write a logical expression. The information encoded in the fixed resource invariants is very useful for identifying unsolvable goal sets without attempting to plan for them. For example, in the ICPARC version of the three-blocks Blocks world (Liatsos & Richards, 1997), in which there are only three table positions, there must always be exactly three clear surfaces. Any goal specifying more than three *clear* relationships can be identified as unachievable from the fixed-resource invariants for that domain. The fixed-resource and uniqueness invariants produced by TIM can be seen as providing a form of *multi-mutex* relations, in contrast to the binary mutex relations inferred during the construction of the plan graph in Graphplan-based planners (Blum & Furst, 1995). Binary mutex relations indicate that two actions or facts are mutually incompatible, whilst multi-mutex relations indicate that larger groups of actions or facts are collectively incompatible. Binary mutex relations, preventing a fact that can be true of only one object from holding of two different objects simultaneously, can be extracted from the identity invariants that TIM infers. Multi-mutex relations are more powerful than binary ones. STAN can detect unsolvable goal-sets by using the fixed-resource and uniqueness invariants even when the binary mutex relations at the corresponding level do not indicate that any problem exists.

To infer these invariants we examine the predicates in the language to see whether they are exchanged on the add and delete lists of the operator schemas. If a predicate is exchanged equally in all schemas (it always appears the same number of times on the add list as on the delete list of a schema) then the predicate corresponds to a fixed resource. If a single schema upsets this balance then the predicate is *not* treated as fixed. Given a fixed resource predicate, it can be inferred that there can never be more combinations of objects satisfying that predicate than there are in the initial state. Because of the slightly odd encoding of the rocket world considered in this paper, only *location* is a fixed resource. *at* is not fixed because it is not equally exchanged in the *load* schema. Examples of fixed resource invariants inferred from various standard domains are provided in Appendix C.

There are certain circumstances under which it is necessary to infer the weaker invariant that

$$|\{x : P(x)\}| \leq k$$

for some positive integer $k$. If $P$ holds of multiple objects in the initial state then it is possible for subsequent state transformations, or attribute acquisitions, to result in states in which two or more instances of $P$ collapse into one. If $P$ holds multiply often in the initial state (or in any other reachable state) then it is necessary to build the invariant using $\leq$ instead of $=$. If $P$ is state-valued, and multiple instances never occur in any state in its property space, then it is safe to assert equality in the construction of the invariant.

Automatic inference of the first three kinds of invariants relies on the construction of the property spaces as discussed in Section 2.4. As has been discussed, the distinction between state and attribute spaces is critical for the inference of correct invariants. However, using just the techniques described so far, TIM would lose information from which it could construct useful invariants. To give an example of how this could occur we now consider the following simple encoding of the standard Blocks world:

**move(X,Y,Z)**

| | | | |
|---|---|---|---|
| Pre: | on(X,Y), | clear(X), | clear(Z) |
| Add: | on(X,Z), | clear(Y), | clear(table) |
| Del: | on(X,Y), | clear(Z) | |

In this operator, used by Bundy *et al.* (1980), the add list element *clear(table)* makes reference to a constant. If the operator schema were to be submitted to our analysis in its current form no PRS would be built for the constant, so the rules that would be constructed, and hence the state and attribute spaces constructed, would fail to record the fact that every application of **move** results in a state in which the table is clear. The resulting analysis would result in incorrect invariants and types. Grant (1996) identifies this version of the **move** operator as flawed, because of the need to maintain state correctness by the addition of the invariant $clear(table)$ to the add list. However, we can analyse this schema correctly if we first abstract it to remove the constant, yielding the following new schema:

**move(X,Y,Z,T)**

| | | | |
|---|---|---|---|
| Pre: | on(X,Y), | clear(X), | clear(Z), table(T) |
| Add: | on(X,Z), | clear(Y), | clear(T) |
| Del: | on(X,Y), | clear(Z) | |

Now, given an initial state in which $blockC$ is on $blockA$ and $blockB$ is on the table, we add the proposition $table(table)$ (so that the new precondition can be satisfied) and the property and attribute spaces that are constructed are as follows:

| | | |
|---|---|---|
| $\{on_1\}$ | $clear_1 \Rightarrow on_1 \rightarrow on_1$ | $\{blockA, blockB, blockC\}$ |
| | | $[on_1]$ |
| $\{on_2, clear_1\}$ | $on_2 \rightarrow clear_1,\ clear_1 \rightarrow on_2,$ | $\{blockA, blockB, blockC, table\}$ |
| | $table_1 \Rightarrow null \rightarrow clear_1$ | |

The second of these is an attribute space, so our invariant extraction algorithm is not applied to it. Consequently, the only invariants we can infer are those that characterize the positions of blocks (every block is on exactly one surface). This is a pity, as there is information available in the attribute space that could yield useful extra invariants. In particular, we would like to infer the invariant that every *block* can be either clear or have something on it, but it cannot be both clear *and* have something on it. The reason we cannot infer this as an invariant is because it would be asserted to hold for every object in the attribute space, including the table, even though it is not actually true of the table (the table can have things on it and still be clear).

#### 2.7.1 Sub-space Analysis on Property and Attribute Spaces

The solution to the problem of loss of invariants is to decompose any property or attribute space that contains $k > 1$ object types into $k$ *sub-spaces*. A property sub-space is structurally identical to a property space. Attribute sub-spaces are identified but not used, as no invariants can be obtained from them. *Property* sub-spaces can be obtained by analysis on *attribute* spaces, as the following example will show. The reason for distinguishing sub-spaces from property and attribute spaces is that the properties are not partitioned in sub-spaces as they are in the property and attribute spaces. The original property or attribute space is not discarded and the sub-spaces are not used for determining the types of objects. The only role of the sub-space analysis is to enable the construction of additional invariants.

We now consider the Blocks domain described in the previous section as an example of the benefits of sub-space analysis. At the point of invariant construction the types of the domain objects have been identified by their property and attribute space membership, so *table* is already known to be of a different type to that of the blocks. This is because *table* is not a member of the property space for $on_1$. Therefore, two sub-spaces can be constructed from the attribute space, one for the type [11], of blocks, and one for the type [01], of tables. No sub-spaces can be constructed from the property space because it contains only one type of object. The rules associated with the sub-spaces will be all of the rules from the original attribute space that are enabled by objects of the appropriate type. The second of the two sub-spaces is an attribute sub-space because of the inclusion of the increasing attribute transition rule. At this stage the two sub-spaces are as follows:

| | | |
|---|---|---|
| $\{on_2, clear_1\}$ | $on_2 \rightarrow clear_1$, $clear_1 \rightarrow on_2$ | $\{blockA, blockB, blockC\}$ |
| $\{on_2, clear_1\}$ | $table_1 \Rightarrow null \rightarrow clear_1$, $on_2 \rightarrow clear_1$, $clear_1 \rightarrow on_2$ | $\{table\}$ |

The attribute sub-space will not be used for invariant construction because it contains an attribute transition rule and would result in incorrect invariants (as is the case for attribute spaces), so there is nothing to be gained from developing it further. However, the state sub-space is now completed by the addition of the states associated with the objects in the space, both in the initial state and by extension. The resulting sub-spaces are:

| | | |
|---|---|---|
| $\{on_2, clear_1\}$ | $on_2 \rightarrow clear_1$, $clear_1 \rightarrow on_2$ | $\{blockA, blockB, blockC\}$ $[on_2]$, $[clear_1]$ |
| $\{on_2, clear_1\}$ | $table_1 \Rightarrow null \rightarrow clear_1$, $on_2 \rightarrow clear_1$, $clear_1 \rightarrow on_2$ | $\{table\}$ |

From the new state sub-space we can infer the following invariants, using the type name *Block* to stand for the type vector [11]. We infer the identity invariant:

$$\forall x : Block \cdot (\forall y \cdot \forall z \cdot (on(y, x) \wedge on(z, x) \rightarrow y = z))$$

the state membership invariant:

$$\forall x : Block \cdot (\exists y : Block \cdot on(y, x) \vee clear(x))$$

and the unique state invariant:

$$\forall x : Block \cdot \neg(\exists y : Block \cdot (on(y, x) \wedge clear(x)))$$

Although there is an additional invariant, that the table is always clear, we cannot infer this at present.

### 2.8 The Problem of Mixed Spaces

It can happen that the encoding of a domain conceals the presence of attributes within schemas until the point at which property space extension occurs. This can prevent the property space extension process from terminating. For example, a simple lightswitch domain contains the following two schemas:

**switchon(X)**
Pre: off(X)
Add: on(X), touched(X)
Del: off(X)

**switchoff(X)**
Pre: on(X)
Add: off(X), touched(X)
Del: on(X)

and an initial state in which *switchA* is *on*. Two PRSs are constructed:

$$\begin{array}{ll} precs: & off_1 \\ deleted_precs: & off_1 \\ add_elements: & on_1, touched_1 \end{array}$$

$$\begin{array}{ll} precs: & on_1 \\ deleted_precs: & on_1 \\ add_elements: & off_1, touched_1 \end{array}$$

giving rise to two rules:

$$off_1 \rightarrow on_1, touched_1$$

and

$$on_1 \rightarrow off_1, touched_1$$

Uniting then seeds one property space containing all three properties. After addition of the rules the property space is as follows:

$$\begin{array}{lll} \{on_1, off_1, touched_1\} & off_1 \rightarrow on_1, touched_1, & \{switchA\} \\ & on_1 \rightarrow off_1, touched_1 & [on_1] \end{array}$$

It is at the point of extension of the space that the problem arises. The following states are added: $[\mathit{off}_1, \mathit{touched}_1]$, $[on_1, \mathit{touched}_1, \mathit{touched}_1]$, $[\mathit{off}_1, \mathit{touched}_1, \mathit{touched}_1, \mathit{touched}_1]$ and so on. We cannot simply avoid adding properties that are already in the state being extended because the two, apparently identical, properties might in general refer to different arguments.

The problem here is due to the fact that $\mathit{touched}_1$ is actually an increasing attribute but this does not become apparent in the PRSs. The consequence is that *mixed spaces* are constructed. A mixed space is a property space containing hidden attributes. TIM detects hidden attributes by checking, on extension, that no new state contains a state already generated from the same initial state starting point. Thus, on extension of the mixed space above, TIM would detect the hidden attribute when the state $[on_1, \mathit{touched}_1, \mathit{touched}_1]$ is constructed, because this state contains the state $[on_1]$ that initiated this extension.

Having detected the hidden attribute there are two possibilities: either TIM can convert the mixed space into an attribute space, in which case no invariants will be constructed, or it can attempt to identify the attribute and split the mixed space into an attribute space and a property space containing the state-valued components of the mixed space. We take this option and split the state. This allows us to infer invariants concerning the state-valued properties.

TIM takes the difference between the including and included states and, for each distinct property in the difference, processes the rules by *cutting* any rule containing that property into two rules, at least one of which will be an attribute rule. The following method is used to cut the rules. In the following, $attr^+$ indicates one or more occurrences of the attribute-valued property and the comma is overloaded to mean both bag conjunction and bag union. If the rule is of the form:

$$\mathit{enablers} \Rightarrow \mathit{start} \rightarrow \mathit{adds},\ \mathit{attr}^+$$

then the two new rules will be of the forms:

$$\mathit{enablers},\ \mathit{start} \Rightarrow \mathit{null} \rightarrow \mathit{attr}^+$$

and

$$\mathit{enablers} \Rightarrow \mathit{start} \rightarrow \mathit{adds}$$

If the rule is of the form:

$$\mathit{enablers} \Rightarrow \mathit{attr}^+,\ \mathit{precs} \rightarrow \mathit{adds}$$

then the two new rules are of the forms:

$$\mathit{enablers}, \mathit{precs} \Rightarrow \mathit{attr} \rightarrow \mathit{null}$$

and

$$\mathit{enablers}, \mathit{attr} \Rightarrow \mathit{precs} \rightarrow \mathit{adds}$$

The rule cutting separates the attribute-valued properties from the state-valued properties. Now pure attribute and property spaces can be constructed. However we do not discard the original mixed space because it has been used in determining the type structure of the domain. Any additional type information that could be extracted from the state and attribute spaces built following this analysis is not currently exploited.

When this analysis is applied to the lightswitch domain, the following new property space and attribute space are built:

| | | |
|---|---|---|
| $\{on_1, off_1\}$ | $off_1 \rightarrow on_1,\ on_1 \rightarrow off_1$ | $\{switchA\}$ |
| | | $[on_1],\ [off_1]$ |
| $\{touched_1\}$ | $off_1 \Rightarrow null \rightarrow touched_1,\ on_1 \Rightarrow null \rightarrow touched_1$ | $\{switchA\}$ |

Using Lightswitch to stand for the type [11], the following state membership invariant can be constructed from the property space:

$$\forall x : Lightswitch \cdot (on(x) \vee off(x))$$

TIM also constructs the uniqueness invariant:

$$\forall x : Lightswitch \cdot \neg(on(x) \wedge off(x))$$

## 3. Properties of TIM

The correctness of TIM relies on it constructing only necessarily true invariants. The demonstration that only true invariants are constructed guarantees the construction of an adequately discriminating type structure. We cannot guarantee against under-discrimination but we argue that over-discrimination does not occur in the type structures generated by TIM. These properties were defined in Section 2.1.

Over-discrimination would be the result of distinguishing functionally identical objects at the type level. This would occur if TIM placed objects that participate in identical state transitions in different property spaces but, because of the underlying partitioning of properties between property spaces, this cannot happen. Further, membership of different property spaces requires that there be distinguishing state transformations, which there are not in functionally identical objects. Flawed assignment (assigning an object to a property space *without* its corresponding state transformations), should simply be seen as erroneous, rather than as over-discrimination. The possibility of this occurring can be excluded because property and attribute space construction and extension are shown to be correct in Section 3.1.

A failure to detect type differences (under-discrimination) in the domain will result in weak invariants, and over-discrimination, if it could occur, would lead to over-targeted invariants that would still be true, but only for a subset of the objects they ought to cover. Flawed assignment would clearly lead to the construction of false invariants. Under-discrimination, which can arise, therefore affects the completeness of the state-invariant inference procedure. It can also lead to *over*-generalisation of the operators since the types assigned to the operator parameters will be equally under-discriminating. This can enable meaningless instances to be formed, needlessly increasing the size of the search space that must be explored by the planner. This clearly raises efficiency issues but it does not undermine the formal properties of the planner that exploits TIM.

As observed, the consequence of under-discrimination is the construction of weak (but valid) invariants. The following example illustrates how under-discrimination can occur. Given a schema:

**op(X,Y)**

| | |
|---|---|
| Pre: | p(X,Y) |
| Add: | q(X,Y) |
| Del: | p(X,Y) |

and an initial state in which

$$p(a, c), p(b, c), q(b, d)$$

hold, the following two property spaces are constructed:

$$\begin{array}{lll} \{p_1, q_1\} & p_1 \rightarrow q_1 & \{a, b\} \\ & & [p_1], [q_1], [p_1, q_1], [q_1, q_1] \\ \{p_2, q_2\} & p_2 \rightarrow q_2 & \{c, d\} \\ & & [q_2], [p_2, p_2], [q_2, p_2], [q_2, q_2] \end{array}$$

Given these property spaces it is impossible to distinguish $a$ from $b$ or $c$ from $d$, even though analysis of the operator schema and initial state reveal that $a$ is functionally distinct from $b$ and $c$ from $d$. It can be seen that, although $a$ must always exchange a $p_1$ for a $q_1$, $b$ can have both $p_1$ and $q_1$ simultaneously. A similar observation can be made for $c$ and $d$. However, the process by which invariants are constructed cannot gain access to this information. An identity invariant constructed for the first property space is:

$$\forall x : T \cdot \forall y \cdot \forall z \cdot \forall u \cdot (q(x, y) \wedge q(x, z) \wedge q(x, u) \rightarrow y = z \vee y = u \vee z = u)$$

This invariant is weaker than is ideal, because $a$ can participate in only one $q$ relation ($b$ can participate in two simultaneously). A state membership invariant for this property space is:

$$\forall x : T \cdot ((\exists y : T_1 \cdot p(x, y)) \vee \exists y : T_1 \cdot q(x, y))$$

which understates the case for $b$, which can have $p_1$ and $q_1$ simultaneously. No unique state invariant is constructed for this property space, because $p_1$ and $q_1$ are not mutually exclusive.

### 3.1 Correctness and Completeness of the Transition Rule Construction Phase

The correctness of the algorithm used in TIM depends on two elements. Firstly, the property spaces identified by the algorithm must be correctly populated. That is, no objects should be assigned to property spaces to which they do not belong and every achievable state must be included in the appropriate property space. Secondly, these property spaces must only support the generation of correct invariants. This second element is examined in Section 3.2.

An interesting relationship exists between the states in a property space and the invariants generated from the space. Incorrect invariants will be contructed if a property space is missing achievable states. This is because the state membership invariants assert that each object in the property space must be in one of the states in the property space. If states are missing then this invariant will be false. We now prove that all achievable states will be in the appropriate property space.

**Theorem 1** *Given an initial state, $I$, a collection of operator schemas, $O$, a property space, $P = (Ps, TRs, Ss, Os)$, generated by* TIM *when applied to $I$ and $O$, and any state, $St$, which* *is reachable from $I$ by application of a valid linearised plan formed from ground instances of operator schemas in $O$, then for any $o \in Os$, the $P$-projection of $St$ for $o$, $St_P^o$, is in $Ss$.*

**Proof:**

The proof is by induction on the length of the plan that yields the state $St$. In the base case the plan contains no operator instances so $St = I$. The $P$-projection of $I$ for $o$ is in $Ss$, by definition of the first phase of the property space construction process described in Section 2.4.

Suppose $St$ is generated by a plan of length $k+1$, with last step $a$ and penultimate state *pre-St*. Let the $P$-projection of *pre-St* for $o$ be $\textit{pre-St}_P^o$. By the inductive hypothesis, this state is in $Ss$. If $a$ does not affect the state of $o$, then the $P$-projection of $St$ for $o$ will be $\textit{pre-St}_P^o$, and therefore in $Ss$ trivially. Otherwise, consider the operator schema, $Op \in O$, from which $a$ is formed. As described in Section 2.7, no constants appear in $Op$ and all variables in the body of $Op$ are parameters of $Op$. Let the *initial* collection of PRSs constructed from $Op$, for those parameters instantiated with $o$ in the creation of $a$, be the set $\text{PRS}_1 \ldots \text{PRS}_n$ where every $\text{PRS}_i$ has the form:

$$\begin{array}{ll} \textit{precs}: & P_i \\ \textit{deleted_precs}: & D_i \\ \textit{add_elements}: & A_i \end{array}$$

and the *initial* collection is the collection formed prior to splitting.

For each value of $i$ the *ith* PRS will lead to the construction of $k+1$ transition rules, where $k$ is the size of the bag intersection, $X_i$, of $D_i$ and $A_i$. The $k$ rules will be of the following form:

$$\forall c \in X_i \cdot (P_i \ominus \{c\} \Rightarrow c \rightarrow c)$$

and the remaining rule will be of the form:

$$P_i \ominus (D_i \ominus X_i) \Rightarrow (D_i \ominus X_i) \rightarrow (A_i \ominus X_i)$$

We refer to the latter rule for $\text{PRS}_i$ as the *ith complex rule.* A subset of the $n$ complex rules will contain a property in $Ps$ in either the *start* or the *finish* and will, therefore, be relevant to the transition from *pre-St* to *St*. It can be observed that these $m$ complex rules ($\text{PRS}_1 \ldots \text{PRS}_m$ without loss of generality) must be in $P$ because of the uniting process described in 2.3.

We define $pres(a)_P^o$ to be the $P$-projection of the preconditions of $a$ for $o$. Similarly, $adds(a)_P^o$ and $dels(a)_P^o$ are defined to be the $P$-projections of the add and delete lists respectively. By construction of the PRSs, defined in Section 3.1,

$$pres(a)_P^o = \bigoplus_1^m P_i$$

$$adds(a)_P^o = \bigoplus_1^m A_i$$

$$dels(a)_P^o = \bigoplus_1^m D_i$$

Because of the restriction that delete lists must be a subset of preconditions, and the fact that $a$ is applicable to *pre-St*, it follows that $dels(a)^o_P \sqsubseteq pres(a)^o_P \sqsubseteq \textit{pre-St}^o_P$. Since $\sqsubseteq$ represents bag inclusion it can be seen that all of the separate bags $D_i$ are included in $\textit{pre-St}^o_P$ without overlap.

The extension process involves the iterated application of the rules as explained in Section 2.4 and indicated in the pseudo-code algorithm presented in Appendix B.

For a rule to be applicable to a state its *start* must be included in the state. Therefore the $m$ complex rules are all applicable, regardless of the sequence of application, to $\textit{pre-St}^o_P$. It follows that the state

$$(\textit{pre-St}^o_P \ominus \bigoplus_1^m (D_i \ominus X_i)) \oplus \bigoplus_1^m (A_i \ominus X_i)$$

is generated in the extension process. By definition of $X_i$, and the fact that $D_i \sqsubseteq \textit{pre-St}^o_P$, this state can be written as:

$$(\textit{pre-St}^o_P \ominus \bigoplus_1^m D_i) \oplus \bigoplus A_i$$

which, as observed above, is just:

$$(\textit{pre-St}^o_P \ominus dels(a)^o_P) \oplus adds(a)^o_P$$

which equals $St^o_P$ by the standard semantics of operator application in STRIPS.

□

The proof demonstrates that splitting, discussed in 2.3, does not result in the generation of invalid invariants. However, splitting can compromise the completeness of the invariant-generation process. It can result in the inclusion of unreachable states in property spaces, with the consequence that the identity and state membership invariants that are generated are weaker than would otherwise be the case. This is further discussed in Section 3.2.

We now explain the role of splitting in the PRS construction phases. Each domain object in a STRIPS domain has an associated finite automaton in which the states consist of the properties (for example, $at_1$) it can have, either initially or as a result of the application of an arbitrary length sequence of operators. Objects that can be observed to be of the same type will have identical automata at the property level. The PRSs capture the ways in which operator applications modify the configurations of individual objects and hence provide an encoding of these automata.

The PRSs are built in two phases. In the first phase, all of the parameters in all of the schemas are considered, so all possible object state transitions are captured. However, some of these transitions conceal the functional distinctions inherent in the domain description and would lead to premature amalgamation of property spaces, as was observed in the discussion of the Rocket domain in Section 2.5. In that example it was observed that use of our standard formula for the construction of rules from these PRSs alone would result in the failure to detect the type distinction between *rockets* and *packages*.

The second phase assists the type inference processes in avoiding under-discrimination by distinguishing *enablers* of a state transformation from the properties that are *exchanged* during the transformation. Each PRS characterizing the exchange of $k$ properties is split to form at most $k+1$ new PRSs. The PRSs 4 and 5, given in Section 2.3, show how two PRSs are constructed from a single PRS containing a single exchanged property. This is a simple example, as only one split is required to remove exchanges. In general it might be necessary to split repeatedly until all exchanges are removed, as shown in the example given by PRS 6 in Section 2.3. No non-exchange combinations of the properties in *deleted_precs* and *add_elements* should be considered during splitting. The resulting PRSs lead to the construction of transition rules which allow generic state transformations, such as movement from one location to another, to be separated from the specific nature of the objects that can make those transformations.

It can be observed that the rules that result from the splitting process are more general than the rules that would have been obtained from the PRS prior to splitting. They distinguish more precisely between the properties that take part in state transitions and the properties that simply enable those transitions, allowing finer type distinctions to be inferred on the basis of the functionalities of the objects in the domain. Finer distinctions are made during the process of seeding property and attribute spaces by uniting. This is because uniting merges, into single equivalence classes, all of the properties that appear in both the start and finish of a rule.

We argue that all state transformations are accounted for by the end of this second phase. The result of the second phase is that the automata formed during the first phase are separated into collections of simpler automata where possible, so that no transitions are lost but there is a finer grained encoding of the possible transitions that can be made by objects with appropriate properties. The PRSs constructed in this phase support the construction of rules that allow objects making these transitions to occupy different property spaces. Some of the second phase PRSs may be under-constraining, in the sense that analysis of subsequent schemas might eliminate the possibilities they are keeping open, as in example 2.3.1, but the set of PRSs obtained at the end of the second phase cannot be over-constraining because all of the first phase PRSs are considered for splitting.

A subtlety concerns the consequence, at the type level, of assigning two functionally distinct objects to the same state or attribute space. For example, in example 2.5, *rocket* and *package* are both assigned to the property space for $\{in_1, at_1\}$ and the attribute space for $\{in_2\}$. However, because *rocket* can be *fuelled* or *unfuelled*, and the *package* cannot, there is a distinction between them that emerges in the property and attribute membership vectors associated with the *rocket* and *package* objects. Membership of the additional property space for $\{fuelled_1, unfuelled_1\}$ means that *rocket* is assigned a type that is a sub-type of the type of *package* and the functional distinctness of *rocket* and *package* is recognised. As discussed, there is an oddity in this encoding that results in the *package* being assigned membership of the $\{in_2\}$ attribute space. Furthermore, $at_1$ and $in_1$ were united, with the effect that rockets can make the $at_1 \rightarrow in_1$ transition and can be used to instantiate variables of type *movable object*, even when variables of this type are intended only to be instantiated with the *package*. There is nothing in the domain description to prevent this interpretation. A more conventional encoding of the *load* schema would prevent the *rocket* from being loaded into any other object, and this would cause a refinement in the type structure that would identify *loadable objects*, and would prohibit the use of the *rocket* in forming instances of operators that should be restricted to operating on those objects.

The construction of transition rules follows a simple rule whereby any undeleted preconditions are used to enable a transformation from a state in which the deleted preconditions of a PRS hold to one in which the added elements of the PRS hold. Given the assumption that all deleted atoms in an operator schema must appear as preconditions in that schema, these rules correctly characterize STRIPS-style state transformations. All possible transformations are captured because of the second phase of PRS construction. A complete set of correct transition rules is therefore constructed.

Given the correctness and completeness of the transition rule construction phase, correct *initial* allocation of objects to spaces depends simply on correctly checking membership of the initial properties of the object in the property sets, formed by uniting the rules, that are used to seed the spaces. Extension of the property spaces is done by straightforward application of the transition rules, so all configurations of properties that can be occupied by the objects in the property space will have been added by the end of the extension phase. Extension of the attribute spaces is unproblematic in the cases where no potential enabler is itself an attribute. If one is, then the process by which the attribute space of that enabler is completed could, it appears, initiate a loop in the attribute space extension process. In fact, this does not happen as TIM is able to detect when a loop has occurred and avoid repeatedly iterating over it.

The following example illustrates the problem and the way it is solved in TIM. Suppose we have three attribute spaces:

**Attribute space 1**

$$\{q_1\} \quad p_1 \Rightarrow null \rightarrow q_1 \quad \{a, b\}$$

**Attribute space 2**

$$\{r_1\} \quad q_1 \Rightarrow null \rightarrow r_1 \quad \{c\}$$

**Attribute space 3**

$$\{p_1\} \quad r_1 \Rightarrow null \rightarrow p_1 \quad \{d\}$$

These spaces are extended by the addition of objects that potentiate their increasing rules, as discussed in Section 2.4. No problem arises if the enablers of these rules are states, and not attributes, but in the extension of attribute space 1 above the enabler, $p_1$, is an attribute. The attribute space for $p_1$ has not yet been extended, so it is necessary to complete that space before using it to complete 1. Extension of 3 requires the extension of 2, for the same reason, and that requires the extension of 1 which requires the extension of 3, and so on.

The way TIM avoids re-entering this loop is by marking each space, as it is considered, as having been seen on this iteration. When a marked space is encountered it is not extended but is used as if it is already complete. Then a second iteration is required to extend any spaces that still require completion. Subsequent iterations will be required until the process converges. Our experiments suggest that it is unusual for there to be more than two iterations required. A worst case upper bound is $o * As$, where $o$ is the number of domain constants and $As$ is the number of attribute spaces (which is limited by the number of properties), and hence quadratic in the size of the domain description.

If the extension process starts with attribute space 1, in the above example, attribute space 1 will be marked as having been seen on the first iteration. Tim then goes on to extend space 3 because the extension of space 1 depends upon space 3 being complete. Space 3 is marked as having been seen on this iteration and space 2 is considered. Space 2 is marked and space 1 is revisited. Because space 1 is marked tim infers that a loop has been entered. Its objects are added to space 2 without extension and the objects of space 2 are then added to space 3. Finally, the objects of space 3 can be added to space 1 and the first iteration is complete.

$$\begin{array}{lll} \{q_1\} & p_1 \Rightarrow null \rightarrow q_1 & \{a,b\} \cup \{c,d\} \\ \{r_1\} & q_1 \Rightarrow null \rightarrow r_1 & \{c\} \cup \{a,b\} \\ \{p_1\} & r_1 \Rightarrow null \rightarrow p_1 & \{d\} \cup \{c,a,b\} \end{array}$$

However, space 2 is not yet complete, so a second iteration is required. This iteration starts in the same place as the first and the process is repeated, except that no further iterations will be required in this example.

### 3.2 Correctness of the State Invariants

We now argue for the correctness of the invariant inference procedure by considering each of the four kinds of invariant in turn. The following arguments rely upon correctly distinguishing property spaces from attribute spaces, since the invariant analysis cannot be performed on attribute spaces. The only scope for confusing this distinction is in the extension of mixed spaces, but we extract attributes from mixed spaces by checking for inclusion of existing states in the new states generated during extension. This process was discussed in Section 2.8.

**Definition 16** *Given a property space $P = (Ps, TRs, Ss, Os)$, $Ss$ can be partitioned into three disjoint sets: $Ss_{subs}$ and $Ss_{sups}$ that contain all of the states in $Ss$ that are included (as bags) or that include (as bags), respectively, at least one other state in $Ss$ and $Ss_{ind}$ that contains all of the independent states in $Ss$ that are neither in $Ss_{subs}$ nor in $Ss_{sups}$.*

**Theorem 2** *Given a property space $P = (Ps, TRs, Ss, Os)$, in which the set of states $Ss$ is a union of the three disjoint sets of states $Ss_{ind}$, $Ss_{subs}$ and $Ss_{sups}$, for each object, $o$, in $Os$ the following families of invariants will hold:*

1. *identity invariants;*
2. *state membership invariants;*
3. *unique state invariants.*

*as defined in Section 3.2.*

**Proof:**

We address each kind of invariant in turn. By Theorem 1 every object in $Os$ must be in a state in $Ss$. Furthermore, all states of each object in $Os$, with respect to each property in $Ps$, will be in $Ss$. This follows because the properties are partitioned between the spaces during the seeding process. Therefore, the maximum number of occurrences of a property $p$ in $Ps$, possessed by any object in $Os$ in any state of the world, will be bounded by the maximum number of instances of that property in any state in $Ss$ (these maximum values might not be equal since $Ss$ can contain inaccessible states). The identity invariants simply express this bound on the properties of the objects in $Os$.

Every object in $Os$ must be in a state in $Ss_{ind} \cup Ss_{subs}$. This follows by definition of these sets in Definition 16 and by Theorem 1. The state membership invariants assert that every object in $Os$ must be in at least one of these states, with each disjunct in the invariant corresponding to the assertion of membership of one of these states.

To argue for the correctness of the unique state invariants, we observe that the proposed invariants would only be false if they paired states that were not mutually exclusive. In this case, either the state extension process would have put properties that could be simultaneously held into the same bag, or such properties would be simultaneously held in the initial state and hence would appear in the same bag on initial construction of the property space. In either case, a state will exist in the property space that is a superset of both of the non-exclusive states. However, uniqueness invariants are generated for pairs of states drawn only from $Ss_{ind} \cup Ss_{sups}$ so these non-exclusive pairs of states will not lead to the generation of incorrect invariants.

□

The fixed resource invariants are always associated with a particular predicate. If atoms built with that predicate are balanced on the add and delete lists of all of the operator schemas then the number of occurrences of these atoms in the initial state is *fixed* over all subsequent states. This is what the invariant expresses. An invariant is constructed for every predicate that forms balanced atoms.

Since no new techniques are required to infer invariants from sub-spaces, no further argument is required to support correctness of the invariants formed following sub-space analysis.

Although Theorem 2 demonstrates the correctness of the invariants inferred by TIM it is possible for weak invariants to be inferred from the presence of unreachable states in $Ss$. Weak identity invariants are inferred if an unreachable state is generated, during extension, containing more instances of a property than are contained in any reachable state. When this happens an identity invariant will be generated that is weaker than would be ideal, but is still valid. Further, if a property space contains unreachable states they will cause the inclusion of additional false disjuncts in the state membership invariants, but since these false disjuncts will not exclude satisfying assignments their presence will not invalidate the invariants. Unreachable states cause additional tautologous uniqueness invariants to be generated but do not affect the strength of the invariants that refer only to reachable states. Clearly we cannot hope to identify all of the unreachable states, as such an analysis would be as hard as planning itself.

Because no invariants are generated for attribute spaces TIM cannot be claimed to be complete. Sub-space analysis rectifies this to some extent by identifying property spaces that exist within attribute spaces and allowing further invariants to be generated. This analysis could be further refined.

### 3.3 Effects of TIM on the Properties of the Planner

Tim is itself sound, so no planner that uses tim is in danger of losing soundness as a result. Tim is certainly not complete for all domain axioms because there are invariant properties of other kinds that cannot be extracted by the current version. For example, Kautz and Selman (1998) identify *optimality conditions* and *simplifying assumptions* amongst the different kinds of axioms that might be inferred from a domain. An optimality condition in the Logistics domain might be: a package should not be returned to a location it has been removed from. A simplifying assumption in the same domain might be: once a truck is loaded it should immediately move (assuming all necessary loads can be done in parallel). These constraints require a deeper analysis of the domain than is currently performed by tim, but we intend to characterise them and infer them in our future work.

We cannot guarantee that the type structure inferred by tim is always fully discriminating, although we do guarantee that it is not over-discriminating. However, failure on tim's part to infer all of the structure that is there to be inferred does not impact on the completeness of a planner using tim because, in these cases, tim will return an unstructured domain and the planner can therefore default to reasoning with the unstructured domain when necessary.

## 4. Experimental Results

An examination of tim's performance can be carried out on several dimensions. We consider three specific dimensions here: the viability of the analysis on typical benchmark domains; the scalability of the analysis and the utility of performing the analysis prior to planning. Its general performance on standard benchmark problems provides an indication of the scale of the overhead involved in using tim as a preprocessing tool. All experiments were performed under Linux on a 300MHz PC with 128 Mb of RAM. Figure 3 shows that, even on large problem instances, the overhead is entirely acceptable. All of the Mystery problems listed in this table are very large (involving initial states containing hundreds of facts) and could not be solved by stan, ipp (Koehler, Nebel, & Dimopoulos, 1997) or Blackbox (Kautz & Selman, 1998) in the aips-98 competition. The nature of the Mystery domain is described in Appendix C. This emphasises the relative costs of the preprocessing and planning efforts.

The selection of problems used to construct table 3 is justified as follows. In the Blocks world we have used a representative example from each of three encodings supplied in the pddl release. These are: the simple encoding (prob12), the att encoding (prob18) and the snlp encoding (prob23). The Hanoi set contains a collection of reasonably sized problems. A representative group of relatively large Mystery instances was chosen from the pddl release. The two Tyre world instances are the only two strips instances available in the release. The three Logistics problems are the three largest for the simple strips encoding included in the pddl release.

The second dimension is scalability of the analysis. An analytic examination of the algorithm can determine an upper bound on performance that is polynomial in all of the key domain and problem components, including number of operator schemas, number of literals in operators, numbers of objects and facts in the initial state and the number and arities of predicates in the language. Figure 4 shows that the performance of tim is roughly quadratic in the size of the problem specification. In the graph, size is crudely equated with the number of characters in the specification file. This graph was constructed by running TIM on all of the STRIPS Mystery domain problems in the PDDL release. The increasing sizes of the problem specifications reflect increases in any and all of the various categories of objects in the domain and corresponding facts to describe their initial states.

| *Domain and problem* | | *Parse time* | *Analysis time* | *Output time* | *Total* |
|---|---|---|---|---|---|
| Blocks | prob12.pddl | 2 | 0 | 2 | 5 |
| | prob18.pddl | 3 | 1 | 2 | 7 |
| | prob23.pddl | 2 | 1 | 1 | 5 |
| Hanoi | 3-disc | 2 | 1 | 4 | 7 |
| | 4-disc | 2 | 1 | 4 | 7 |
| | 5-disc | 3 | 1 | 4 | 8 |
| | 6-disc | 3 | 1 | 4 | 9 |
| | 7-disc | 4 | 2 | 4 | 11 |
| Mystery | prob060.pddl | 17 | 15 | 9 | 43 |
| | prob061.pddl | 48 | 82 | 29 | 160 |
| | prob062.pddl | 26 | 37 | 10 | 74 |
| | prob063.pddl | 11 | 7 | 8 | 27 |
| | prob064.pddl | 21 | 21 | 10 | 52 |
| Tyre-World | prob01.pddl | 5 | 2 | 28 | 36 |
| | prob02.pddl | 6 | 2 | 28 | 37 |
| Logistics | prob04.pddl | 4 | 2 | 5 | 12 |
| | prob05.pddl | 4 | 2 | 6 | 12 |
| | prob06.pddl | 4 | 2 | 6 | 13 |

Figure 3: Table showing TIM's performance in milliseconds on standard domains and problems. All timings are elapsed times and minor discrepancies in totals arise from rounding.

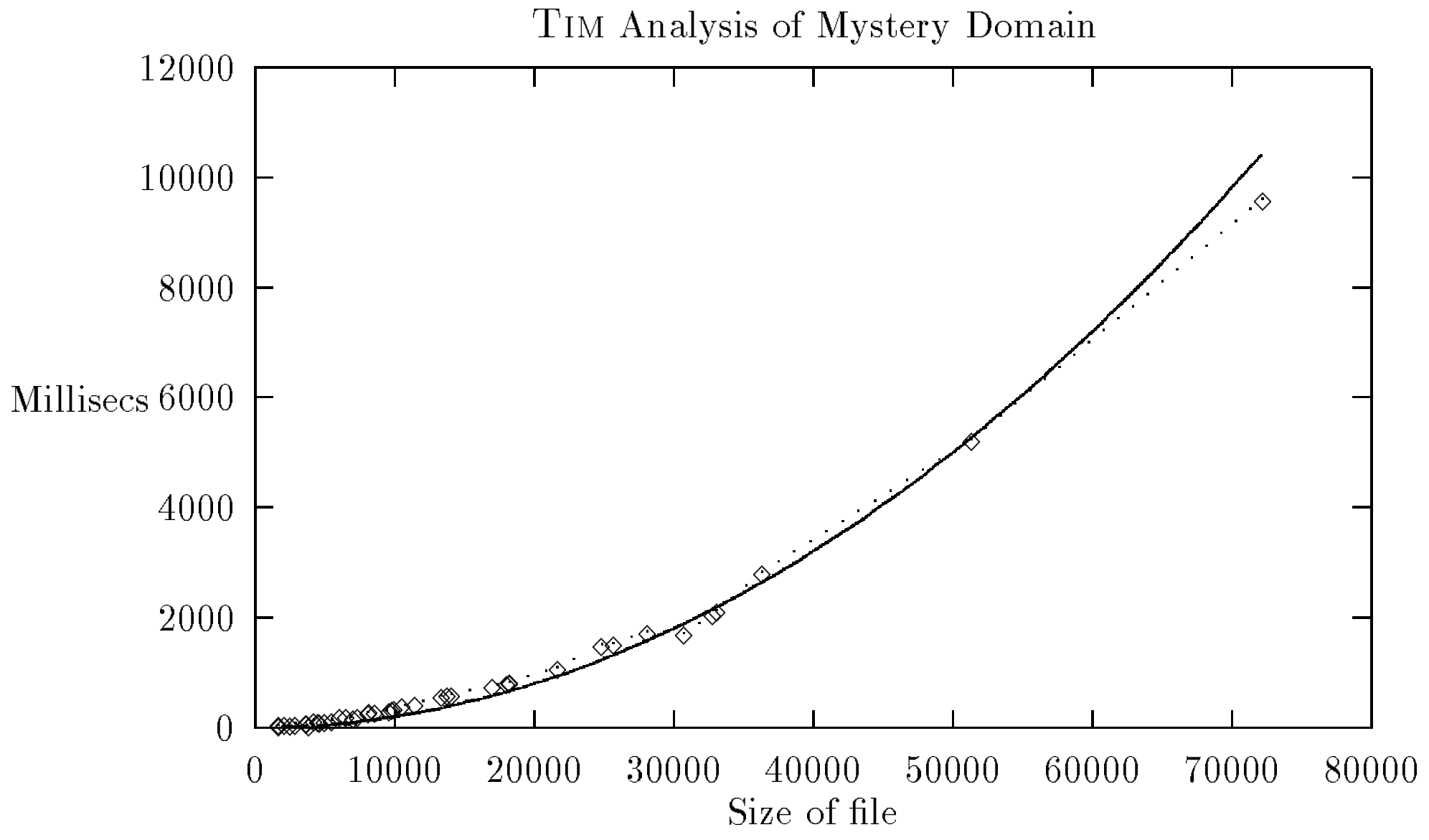

Figure 4: Graph showing TIM's performance on Mystery problems, plotting time against size (in characters) of problem file. The solid line is a plot of a quadratic function.

Figure 5 shows the effect on TIM's performance as the number of operator schemas increases. This graph was constructed using an artificial domain in which each new operator causes two new state transitions described by two new literals. Thus, both number of operators and number of properties is increasing whilst the number of objects stays constant. The domain is described in detail in Appendix E. The graph indicates the linear growth of cost of analysis.

The final dimension for evaluating TIM is the effect of exploitation of its output by a planner. Gerevini and Schubert (1998) and Kautz and Selman (1998) provide convincing evidence supporting the powerful role of state invariants in enhancing the performance of SAT-based planning. In Figure 6 we demonstrate the power of inferred types by showing the advantage that STAN with TIM obtains over STAN without TIM on untyped Rocket domain problems. Figure 6 shows the effect on performance of increasing the number of packages to be transported. The time taken by STAN with TIM grows linearly, whilst STAN without TIM follows a cubic curve. If there are $p$ packages in a problem instance then STAN with TIM constructs $4(p+1)$ operator instances while STAN without TIM constructs $(p+3)^2(p+5)+2p$ instances. This demonstrates that type information is the most significant factor in the advantage depicted in the graph. Figure 7 demonstrates that a similar improvement is obtained in the Logistics domain. In this graph a series of sub-problems were considered in which each sub-problem involves the independent transportation of a single package between two cities.

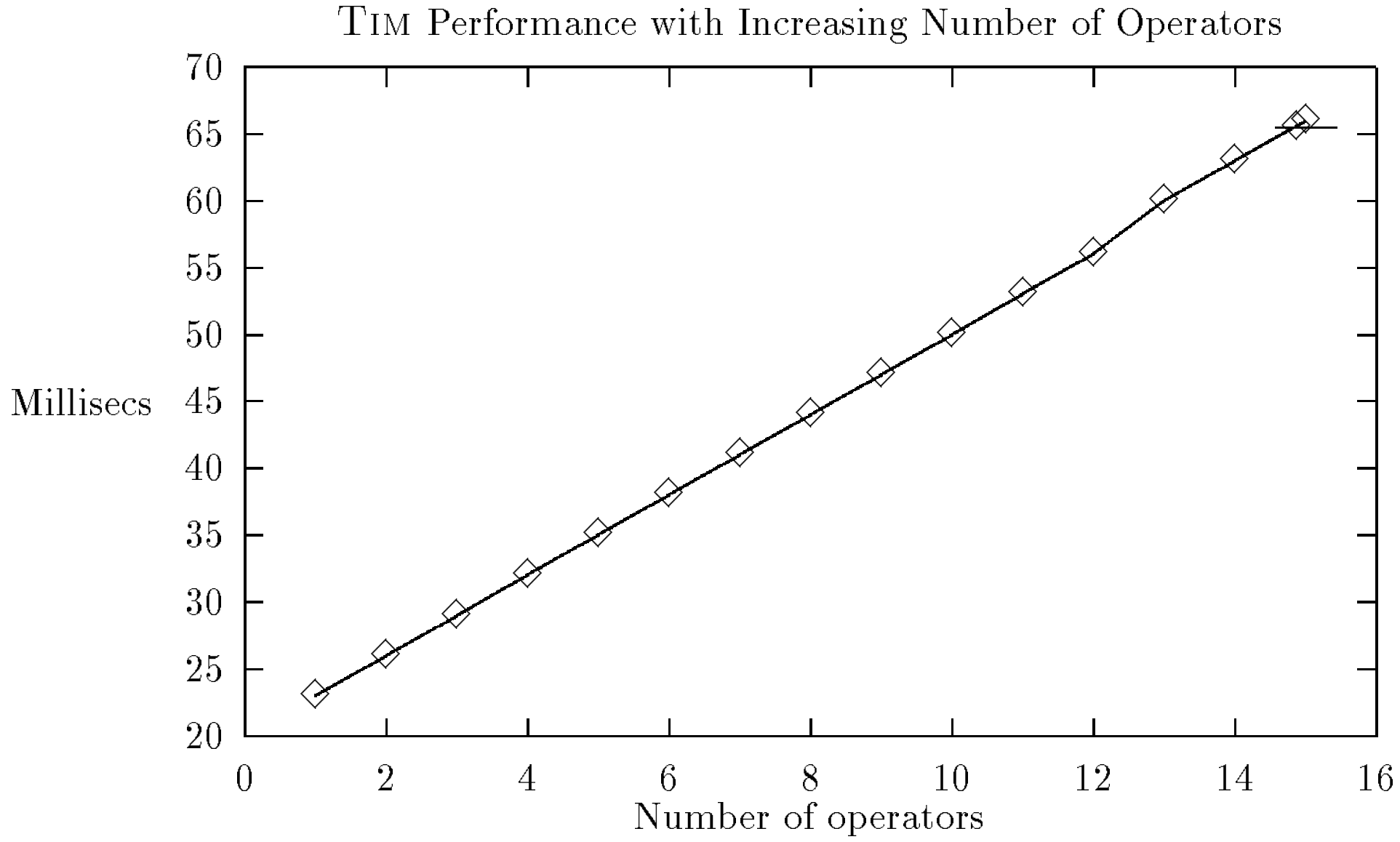

Figure 5: Graph showing the consequences of increasing the number of schemas and inferrable property spaces.

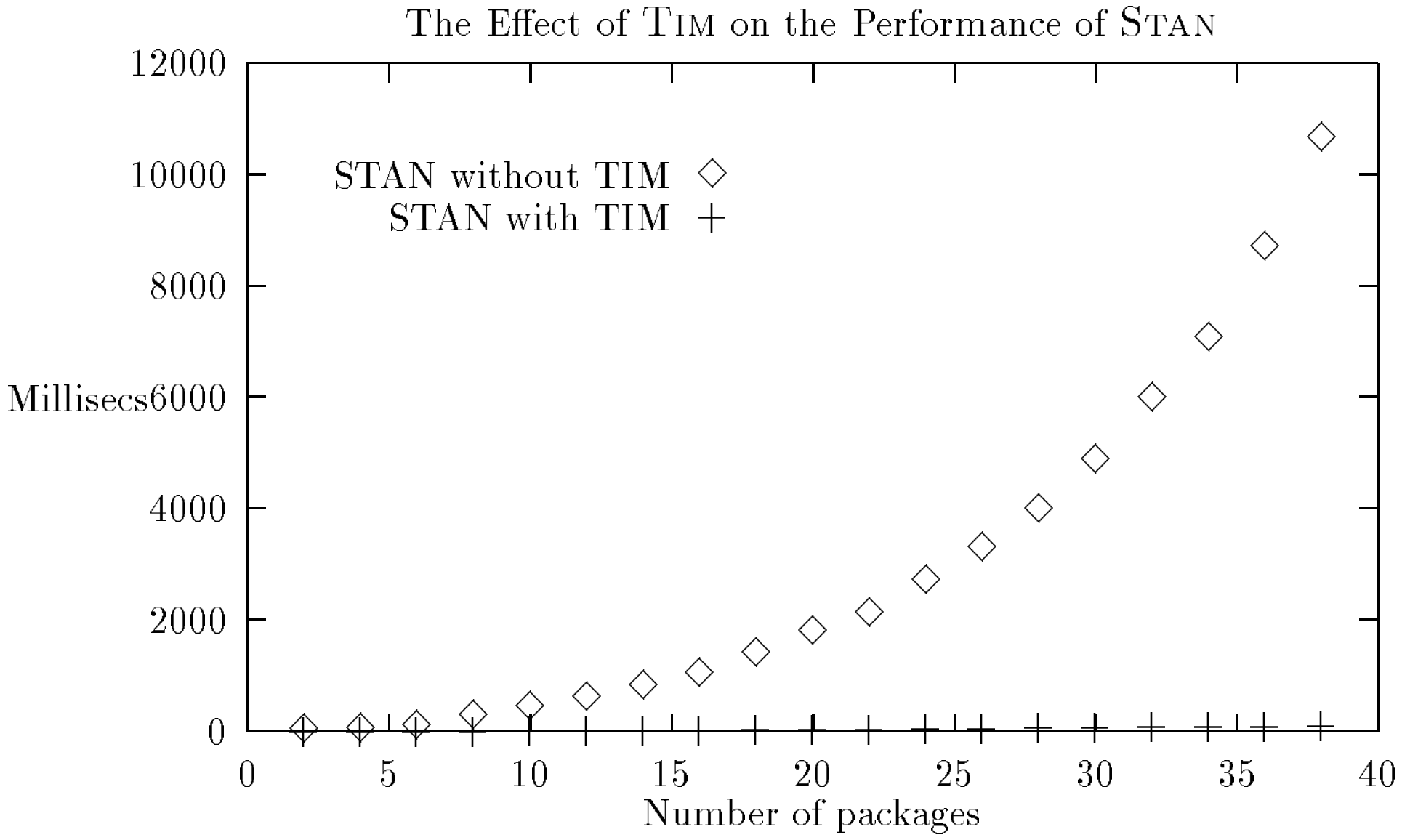

Figure 6: Graph showing comparison between STAN with and STAN without TIM on Rocket domain problems generated from the Rocket domain provided in Appendix D.

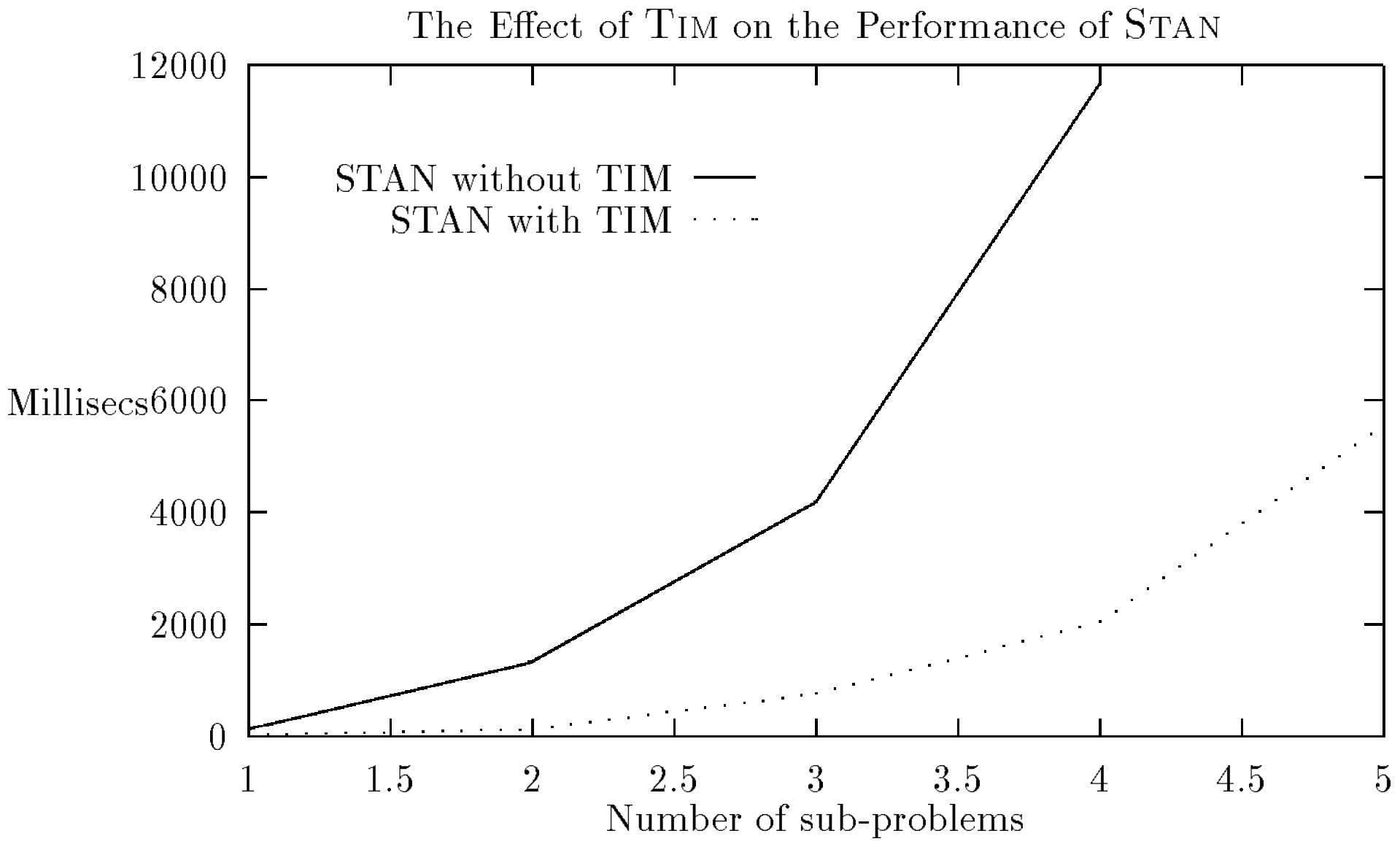

Figure 7: Graph showing comparison between STAN with and STAN without TIM on Logistics domain problems.

In very simple domains, the overhead of carrying out this analysis can outweigh the advantages offered. For example, in the Movie domain used in the competition STAN gained no benefits from using TIM but paid the overhead to the detriment of its performance on instances from that domain. However, in general we observe that the benefits of this analysis increase with the increasing complexity of domains.

## 5. Related Work

Although the importance of state invariants for efficient planning has been observed there has been relatively little work on *automatic* inference of invariants. The published work that most closely resembles the research described in this paper is the state constraint inference system DISCOPLAN, of Gerevini and Schubert (1998). Discoplan enables the inference of *sv-constraints* that correspond to a subset of our identity invariants. The reason that DISCOPLAN is restricted to a subset is that it generates sv-constraints only for pairs of literals (one on the addlist of a schema and the other on the delete list) in which the arguments vary in only one place. Tim can infer identity invariants in which *vectors* of arguments vary, as shown in Section 2.7. Discoplan cannot currently infer all singly varying constraints (although the techniques described by Gerevini and Schubert (1996a) are not yet fully implemented in DISCOPLAN). For example, DISCOPLAN cannot infer that all blocks can only be *on* one surface, in its analysis of the Blocks world domain cited in the paper. TIM can infer these invariants from its sub-space analysis.

Gerevini and Schubert (1996a, 1996b) have also examined the potential for inferring parameter domains that are similar to the operator parameter types inferred by TIM. Their domains are inferred by an iterative process of accretion which is similar to the attribute space extension process of TIM. However, the accretion process they describe is synthetic, in that the parameter domains are synthesised directly from the operator descriptions and initial state. TIM is an analytic system that constructs its types from an analysis of the functional properties of the domain objects. This analytic approach provides a rich information source from which other structures, including the domain invariants, can be derived.

Some of the *implicative* constraints inferred by DISCOPLAN correspond to an implicit type assignment and would arise in the type structure built by TIM. A further implicative constraint generated by DISCOPLAN refers to the separation of functional roles of objects. In particular, the irreflexivity of *on*, as in:

$$\forall x \cdot \forall y \cdot (on(x, y) \rightarrow \neg(x = y))$$

can be captured using this kind of constraint. TIM cannot currently infer these invariants. Because TIM uses an analysis based on the state view of objects in the domain it is able to generate a broader collection of invariants, including state membership and unique state invariants currently not produced by DISCOPLAN.

Although DISCOPLAN can deal with negative preconditions and TIM cannot yet manage them, the invariants they produce overall are currently less powerful than those inferred by TIM.

Apart from the work of Gerevini and Schubert, there is some older work on the inference of invariants which also relies on the generation of candidate invariants which are then confirmed by an inductive process against the domain operators. Two examples are the work of Kelleher and Cohn (1992) and Morris and Feldman (1989). The former work concentrates on identifying *directed mutual persistence* relations, which hold between pairs of facts in a domain when, once both are established, the second continues to hold while the first does. The use of these relations leads to the inference of a collection of constraints which fall into the uniqueness invariants inferred by TIM. In the work described in (Morris & Feldman, 1989) the authors build invariants by using *truth counts* which are counts of the number of propositions from particular identified sets which must be true in any state of the domain. Sets for which this count is 1 can then be used to build invariants which are a subset of the state membership and uniqueness invariants. The authors describe methods for attempting to identify the sets of facts from which to work. This work, in common with that of Kelleher and Cohn and of Gerevini and Schubert, builds invariants by first hypothesising a possible seed for the invariants and then determining their validity by analysing the effects of the operators on these seeds. In contrast to this *generate-and-test* strategy, TIM produces only correct invariants which it infers from a deep, structural analysis of the domain. The inference of invariants does not exhaust the possibilities of this analysis. For example, the type structure is inferred automatically during this analysis, which has been shown to have dramatic potential for the efficiency of planning. The relationship between enablers, and the state transitions they enable, determines an ordering on the satisfaction of goals, which also has significance for efficiency. Further, the state-based view of the behaviour of domain objects would allow the techniques described by McCluskey and Porteous (1997) to be automated.

McCluskey and Porteous (1997) have proposed and explored an *object-centred* approach to planning. This approach is based on the provision, by a domain engineer, of a rich collection of state invariants for object sorts participating in functional relationships in the domain. These invariants are then exploited in a domain compilation phase to facilitate an efficient planning application to that domain. Tim infers precisely the sorts and collections of state invariants that McCluskey and Porteous provide by hand.

Grant (1996) generates state invariants from state descriptions, provided by hand, and then uses these invariants to build operator schemas. His approach is clearly related even though the objectives of his analysis are different. Grant is concerned with the automatic synthesis of domain descriptions from a rich requirements specification provided by an expert user. Our concern is with reverse-engineering a domain description to obtain the information that can help increase the efficiency of planners applied to that domain. Although the primary objectives in the use of tim are to enhance the performance of planning within a domain, tim also provides a valuable tool in the construction of domain descriptions by revealing the underlying behaviours that the domain engineer has implicitly imposed, and helping with the debugging of domain descriptions.

## 6. Conclusion

Tim is a planner-independent set of techniques for identifying the underlying structure of a domain, revealing its type structure and a collection of four different kinds of invariant conditions. One important application of these techniques is as a domain debugging aid during the construction of large and complex domains. Using tim has revealed many anomalies in domains encoded by us and by others, and has greatly assisted us in understanding stan's performance on many domains and problems. Another important application is in increasing the efficiency of planners by making explicit to the planner information about the domain that it would otherwise have to infer, from the domain representation, during planning.

Tim generates a rich collection of invariants containing many that are not inferrable by related systems, as discussed in the previous section. The results presented by Gerevini and Schubert (1998) suggest that a marked improvement can be obtained from the use of invariants in the performance of planners based on SAT-solving techniques. No analysis has yet been done to determine what advantages might be obtainable by using invariants in planners based on other architectures. Stan does not yet exploit all of the invariants produced by tim during planning. It uses the type structure and the fixed resource invariants and we are currently developing an extension of stan that will fully exploit the other kinds of invariant. We expect to be able to use the uniqueness and identity invariants to shortcut the effort involved in deducing a significant subset of the necessary mutex relations during graph construction.

The analysis performed by tim is efficient, growing more slowly than a quadratic function of the size of the initial state being analysed. Our empirical analysis does not consider the effect on tim's performance of increasing numbers of operator schemas. However, the argument presented in Section 4 shows that tim's analysis grows linearly with the number of operator schemas, linearly with the number of domain constants and linearly with the size of the initial state. There are other factors to take into account, but this confirms a polynomial performance as the size (and related structure) of the domain increases.

The type analysis performed by TIM differs, in some important respects, from the various forms of type analysis performed during the compilation of programs written in strongly typed languages. In the latter context the type-correctness of a program is judged with respect to an imposed context of basic types. TIM infers the basic types from the domain description so it is impossible for a domain specification *not* to be well-typed. Consequently we do not attempt to type-check domain descriptions using TIM. This is a direction in which we hope to move in the near future, because type-checking will enable some unsolvable problems to be detected as unsolvable statically rather than at planning time. We currently focus only on type *inference* and the exploitation of the inferred type structure in the management of the search space of the planner.

## 7. Acknowledgements

We would like to thank Alfonso Gerevini, Gerry Kelleher and the anonymous referees for useful discussions and helpful comments on earlier drafts of this paper.

## Appendix A. FTP and Web Sites

The AIPS-98 Planning Competition FTP site is at:
`http://ftp.cs.yale.edu/pub/mcdermott/aipscomp-results.html`.
Our web site, on which STAN and TIM executables can be found, is at:
`http://www.dur.ac.uk/~dcs0www/research/stanstuff/planpage.html`

## Appendix B. The TIM Algorithm

The following is a pseudo-code description of the TIM algorithm.

```
{Construct base PRSs (Section 2.3)}
  Ps := {};
  for each operator schema, O,
      for each variable in O, x,
          construct a PRS for x from O and add to Ps;

{Split PRSs}
  for each PRS in Ps, P,
      if a property, p, appears in P in both the adds and deleted_precs fields
      then split P over p, into P' and Q and replace P with P' and Q in Ps,
              where to split P over p:
                  construct PRS Q with the same precs as P, deleted_precs and adds both set to {p};
                  construct PRS P' from P by removing p from deleted_precs and adds of P;

{Construct transition rules (Section 2.3)}
  Ts := {};
  for each PRS in Ps, P,
      construct a transition rule for P and add to Ts;

{Seed property and attribute spaces (Section 2.3)}
  let each property be initially assigned to a separate equivalence class;
  for each rule, r, in Ts
      merge together (unite) the equivalence classes for all the properties in the start and finish of r;
```

```
   construct a separate space for each equivalence class of properties;

{Assign transition rules (Section 2.4)}
   for each rule, r, in Ts
          place r in the space associated with the equivalence class containing the properties
                 in the start (and finish) of r, s;
          if r is an increasing or decreasing rule
          then  mark s as an attribute space;

{Analyse initial state (Section 2.4)}
   for each object, o, in the domain
          identify the bag of initial properties of o, I(o);
          for each space, s,
                 construct the bag of properties from I(o) which belong to the equivalence class
                        associated with s, b;
                 if b is non-empty
                 then add o to the space s;
                        if s is not an attribute space
                        then   add b as a state in s;

{Extend property spaces (Section 2.4)}
   for each property space, p,
          while there is an unextended state in p, s,
                 mark s as extended;
                 newgen := {};
                 for each rule in p, r,
                        if the start of r is included in s
                        then   add the state snew = (s ominus start oplus end) to newgen;
                               if snewis a superset of any state in newgen
                               then   mark p is an attribute space and exit the analysis of p;
                 add newgen to the states in p;

{Extend attribute spaces (Section 2.4)}
   changes := TRUE;
   while changes,
          changes := FALSE;
          for each unmarked attribute space, a,
                 extend a where to extend a:
                               mark a;
                               for each rule in a, r,
                                      for each property in enablers of r, p,
                                             if p's equivalence class is associated with an unmarked attribute space, a',
                                             then  extend a';
                               add all objects that appear in every space associated with an enabling property for r to a;
                               if objects are added
                               then   changes := TRUE;

{Identify types (Section 2.6)}
   for each object in the domain, o,
          identify the pattern of membership of spaces for o, tt;
          associate the type pattern, tt, with o;
   for each operator schema, O,
          for each argument of O, x,
                 identify the pattern of membership of spaces for x implied by the properties of x in the
                        preconditions of O, tt;
                 associate type pattern, tt, with x in O;

{Construct invariants (Section 2.7)}
   for each property space, P,
          for each property in P, p,
                 construct an identity invariant for p;
          construct a state membership invariant for P;
          construct a uniqueness invariant for P;
```

## Appendix C. Example Output

The following output was produced by TIM and can be found, along with other examples, on the STAN webpage. These examples show the details of the analysis performed on each of three domains: a Flat-tyre domain, a Mystery domain and a Logistics domain. The analysis is done with respect to an initial state and a set of operator schemas. The operator schemas used in these three domains are those provided with the PDDL STRIPS releases for these domains. The initial states were taken from the PDDL release. The PDDL release can be found at http://www.cs.yale.edu/HTML/YALE/CS/HyPlans/mcdermott.html.

### C.1 The Tyre World

```
TIM: Type Inference Mechanism - support for STAN: State Analysis Planner

D. Long and M. Fox, University of Durham


Reading domain file: domain01.pddl
Reading problem file: prob01.pddl

TIM: Domain analysis complete for flat-tire-strips

TIM: TYPES:

Type T0 = {wrench}
Type T1 = {wheel2}
Type T2 = {wheel1}
Type T3 = {trunk}
Type T4 = {the-hub}
Type T5 = {pump}
Type T6 = {nuts}
Type T7 = {jack}
```

It will be noticed that the two wheels are separated into different types. This is because one wheel is intact and the other is not intact, and there is no operator for repairing wheels that are not intact. The tools have each been given different types. This is because they each appear as constants in different operators and therefore are functionally distinct.

```
TIM: STATE INVARIANTS:

FORALL x:T4. (on-ground(x) OR lifted(x))
FORALL x:T4. NOT (on-ground(x) AND lifted(x))

FORALL x:T3. (closed(x) OR open(x))
FORALL x:T3. NOT (closed(x) AND open(x))
```

```
FORALL x:T1 U T2. (deflated(x) OR inflated(x))
FORALL x:T1 U T2. NOT (deflated(x) AND inflated(x))
```

The invariants for hubs (below) suggest that almost anything could be *on* a hub. Since this is not the case the type structure is under-discriminating. However, the additional invariants drawn from the sub-space analysis provide enough information, in principle, to discriminate more fully between the types. This information is not yet being fully exploited.

```
FORALL x:T4. FORALL y1. FORALL z1. on(y1,x) AND on(z1,x) => y1 = z1
FORALL x:T4. (Exists y1:T0 U T1 U T2 U T5 U T6 U T7. on(y1,x)
OR free(x))
FORALL x:T4. NOT (Exists y1:T0 U T1 U T2 U T5 U T6 U T7. on(y1,x)
AND free(x))

FORALL x:T4. FORALL y1. FORALL z1. tight(y1,x) AND tight(z1,x) => y1 = z1
FORALL x:T4. FORALL y1. FORALL z1. loose(y1,x) AND loose(z1,x) => y1 = z1
FORALL x:T4. ((Exists y1:T0 U T1 U T2 U T5 U T6 U T7. tight(y1,x)
AND fastened(x))
OR (Exists y1:T0 U T1 U T2 U T5 U T6 U T7. loose(y1,x)
AND fastened(x)) OR unfastened(x))
FORALL x:T4. NOT ((Exists y1:T0 U T1 U T2 U T5 U T6 U T7. tight(y1,x)
AND fastened(x))
AND (Exists y1:T0 U T1 U T2 U T5 U T6 U T7. loose(y1,x)
AND fastened(x)))
FORALL x:T4. NOT ((Exists y1:T0 U T1 U T2 U T5 U T6 U T7. tight(y1,x)
AND fastened(x)) AND unfastened(x))
FORALL x:T4. NOT ((Exists y1:T0 U T1 U T2 U T5 U T6 U T7. loose(y1,x)
AND fastened(x)) AND unfastened(x))


TIM: DOMAIN INVARIANTS:

|{x0: container(x0)}| = 1
|{x0: hub(x0)}| = 1
|{x0: intact(x0)}| = 1
|{x0: jack(x0)}| = 1
|{x0: nut(x0)}| = 1
|{x0: pump(x0)}| = 1
|{x0: unlocked(x0)}| = 1
|{x0: wheel(x0)}| = 2
|{x0: wrench(x0)}| = 1


TIM: ATTRIBUTE SPACES:
```

The attribute space for the properties in the first of these groups is subjected to a much more rigorous analysis in the sub-space invariants below.

```
Objects, x, in T0 U T1 U T2 U T5 U T6 U T7 can have property:
Exists y1:T3. in(x,y1);
Exists y1:T4. on(x,y1);
Exists y1:T4. tight(x,y1);
Exists y1:T4. loose(x,y1);
have(x);
Objects, x, in T3 can have property:
Exists y1:T0 U T1 U T2 U T5 U T6 U T7. in(y1,x);
Objects, x, in T3 all have property: container(x);
Objects, x, in T4 all have property: hub(x);
Objects, x, in T1 all have property: intact(x);
Objects, x, in T7 all have property: jack(x);
Objects, x, in T6 all have property: nut(x);
Objects, x, in T5 all have property: pump(x);
Objects, x, in T3 all have property: unlocked(x);
Objects, x, in T1 U T2 all have property: wheel(x);
Objects, x, in T0 all have property: wrench(x);


TIM: OPERATOR PARAMETER RESTRICTIONS:

inflate(x1:T1)
put-on-wheel(x1:T1 U T2,x2:T4)
remove-wheel(x1:T1 U T2,x2:T4)
put-on-nuts(x1:T6,x2:T4)
remove-nuts(x1:T6,x2:T4)
jack-down(x1:T4)
jack-up(x1:T4)
tighten(x1:T6,x2:T4)
loosen(x1:T6,x2:T4)
put-away(x1:T0 U T1 U T2 U T5 U T6 U T7,x2:T3)
fetch(x1:T0 U T1 U T2 U T5 U T6 U T7,x2:T3)
close-container(x1:T3)
open-container(x1:T3)
cuss()


TIM: ADDITIONAL STATE INVARIANTS, USING SUB-SPACE ANALYSIS:
```

We report here only the additional state invariants that add information to the invariants already listed. TIM currently reports invariants that are subsumed by the earlier collection.

It should be observed that the first wheel is intact but the second is not, and this gives rise to the following new invariant for wheels of the second type.

```
FORALL x:T2. (deflated(x))
```

The first attribute space, which contains all objects except the trunk and the hub, is now subjected to sub-space analysis yielding a rich new collection of invariants.

```
FORALL x:T0. FORALL y1. FORALL z1. in(x,y1) AND in(x,z1) => y1 = z1
FORALL x:T0. (Exists y1:T3. in(x,y1) OR have(x))
FORALL x:T0. NOT (Exists y1:T3. in(x,y1) AND have(x))

FORALL x:T1. FORALL y1. FORALL z1. in(x,y1) AND in(x,z1) => y1 = z1
FORALL x:T1. FORALL y1. FORALL z1. on(x,y1) AND on(x,z1) => y1 = z1
FORALL x:T1. (Exists y1:T3. in(x,y1) OR have(x)
OR Exists y1:T4. on(x,y1))
FORALL x:T1. NOT (Exists y1:T3. in(x,y1) AND have(x))
FORALL x:T1. NOT (Exists y1:T3. in(x,y1) AND Exists y1:T4. on(x,y1))
FORALL x:T1. NOT (have(x) AND Exists y1:T4. on(x,y1))

FORALL x:T2. FORALL y1. FORALL z1. in(x,y1) AND in(x,z1) => y1 = z1
FORALL x:T2. FORALL y1. FORALL z1. on(x,y1) AND on(x,z1) => y1 = z1
FORALL x:T2. (Exists y1:T4. on(x,y1) OR have(x)
OR Exists y1:T3. in(x,y1))
FORALL x:T2. NOT (Exists y1:T4. on(x,y1) AND have(x))
FORALL x:T2. NOT (Exists y1:T4. on(x,y1) AND Exists y1:T3. in(x,y1))
FORALL x:T2. NOT (have(x) AND Exists y1:T3. in(x,y1))

FORALL x:T5. FORALL y1. FORALL z1. in(x,y1) AND in(x,z1) => y1 = z1
FORALL x:T5. (Exists y1:T3. in(x,y1) OR have(x))
FORALL x:T5. NOT (Exists y1:T3. in(x,y1) AND have(x))

FORALL x:T6. FORALL y1. FORALL z1. in(x,y1) AND in(x,z1) => y1 = z1
FORALL x:T6. FORALL y1. FORALL z1. tight(x,y1)
AND tight(x,z1) => y1 = z1
FORALL x:T6. FORALL y1. FORALL z1. loose(x,y1)
AND loose(x,z1) => y1 = z1
FORALL x:T6. (Exists y1:T4. tight(x,y1)
OR Exists y1:T4. loose(x,y1)
OR have(x) OR Exists y1:T3. in(x,y1))
FORALL x:T6. NOT (Exists y1:T4. tight(x,y1)
AND Exists y1:T4. loose(x,y1))
FORALL x:T6. NOT (Exists y1:T4. tight(x,y1) AND have(x))
FORALL x:T6. NOT (Exists y1:T4. tight(x,y1)
AND Exists y1:T3. in(x,y1))
FORALL x:T6. NOT (Exists y1:T4. loose(x,y1) AND have(x))
FORALL x:T6. NOT (Exists y1:T4. loose(x,y1)
AND Exists y1:T3. in(x,y1))
FORALL x:T6. NOT (have(x) AND Exists y1:T3. in(x,y1))
```

### C.2 The Mystery Domain

The Mystery domain was devised by Drew McDermott for the AIPS-98 planning competition. His intention was to conceal the structure of the problem domain by employing an obscure encoding of a transportation domain. The code replaces locations with the names of *foods* and the routes between them with *eats* relations. The transports are *pleasures* while cargos are *pains*. Cargos and transports can be at locations, with the at relation encoded as *craves*. A cargo is either at a location or in a transport encoded by the *fears* relation. Transports have restricted capacity encoded by *planets* and consume fuel in travelling between locations. Fuel exists in limited quantities at locations measured by *provinces*. Using TIM we were able to decode the domain and identify the roles played by each of the components of the encoding.

```
TIM: Domain analysis complete for mystery-strips (prob048.pddl)

TIM: TYPES:
```

It should be noted that provinces (types T6, T7 and T8) are divided into three separate types because they form a sequence, defined by the *attacks* relation, in which the first and last have a slightly different functional role to the others. The same is true of the planets (types T1, T2 and T3).

```
Type T0 = {beef,cantelope,chocolate,flounder,guava,mutton,onion,
   pepper,rice,shrimp,sweetroll,tuna,yogurt}
Type T1 = {saturn}
Type T2 = {pluto}
Type T3 = {neptune}
Type T4 = {achievement,lubricity}
Type T5 = {abrasion,anger,angina,boils,depression,grief,hangover,
   laceration}
Type T6 = {alsace,bosnia,guanabara,kentucky}
Type T7 = {goias}
Type T8 = {arizona}


TIM: STATE INVARIANTS:

FORALL x:T4. FORALL y1. FORALL z1. harmony(x,y1)
AND harmony(x,z1) => y1 = z1
FORALL x:T4. (Exists y1:T1 U T2 U T3. harmony(x,y1))

FORALL x:T0. FORALL y1. FORALL z1. locale(x,y1)
AND locale(x,z1) => y1 = z1
FORALL x:T0. (Exists y1:T6 U T7 U T8. locale(x,y1))

FORALL x:T4 U T5. FORALL y1. FORALL z1. fears(x,y1)
```

```
AND fears(x,z1) => y1 = z1
FORALL x:T4 U T5. FORALL y1. FORALL z1. craves(x,y1)
AND craves(x,z1) => y1 = z1
FORALL x:T4 U T5. (Exists y1:T0. craves(x,y1)
OR Exists y1:T4. fears(x,y1))
FORALL x:T4 U T5. NOT (Exists y1:T0. craves(x,y1)
AND Exists y1:T4. fears(x,y1))



TIM: DOMAIN INVARIANTS:

|{(x0,x1): attacks(x0,x1)}| = 5
|{(x0,x1): eats(x0,x1)}| = 36
|{x0: food(x0)}| = 13
|{(x0,x1): harmony(x0,x1)}| = 2
|{(x0,x1): locale(x0,x1)}| = 13
|{(x0,x1): orbits(x0,x1)}| = 2
|{x0: pain(x0)}| = 8
|{x0: planet(x0)}| = 3
|{x0: pleasure(x0)}| = 2
|{x0: province(x0)}| = 6


TIM: ATTRIBUTE SPACES:

Objects, x, in T1 U T2 U T3 can have property:
Exists y1:T4. harmony(y1,x);
Objects, x, in T6 U T7 U T8 can have property:
Exists y1:T0. locale(y1,x);
Objects, x, in T4 can have property:
Exists y1:T4. fears(y1,x);
Objects, x, in T0 can have property:
Exists y1:T4 U T5. craves(y1,x);
Objects, x, in T6 U T7 all have property:
Exists y1:T6 U T8. attacks(x,y1);
Objects, x, in T6 U T8 all have property:
Exists y1:T6 U T7. attacks(y1,x);
Objects, x, in T0 all have property:
Exists y1:T0. eats(x,y1);
Objects, x, in T0 all have property:
Exists y1:T0. eats(y1,x);
Objects, x, in T0 all have property: food(x);
Objects, x, in T2 U T3 all have property:
Exists y1:T1 U T2. orbits(x,y1);
```

```
Objects, x, in T1 U T2 all have property:
Exists y1:T2 U T3. orbits(y1,x);
Objects, x, in T5 all have property: pain(x);
Objects, x, in T1 U T2 U T3 all have property: planet(x);
Objects, x, in T4 all have property: pleasure(x);
Objects, x, in T6 U T7 U T8 all have property: province(x);


TIM: OPERATOR PARAMETER RESTRICTIONS:

succumb(x1:T5,x2:T4)
feast(x1:T4,x2:T0,x3:T0)
overcome(x1:T5,x2:T4)


TIM: ADDITIONAL STATE INVARIANTS, USING SUB-STATE ANALYSIS:
```

These additional invariants show that the transports are always at a location and never loaded into other transports.

```
FORALL x:T4. FORALL y1. FORALL z1. craves(x,y1)
AND craves(x,z1) => y1 = z1
FORALL x:T4. (Exists y1:T0. craves(x,y1))
```

### C.3 The Logistics Domain

```
TIM: Domain analysis complete for logistics-strips (prob05.pddl)

TIM: TYPES:

Type T0 = {bos-truck,la-truck,pgh-truck}
Type T1 = {bos-po,la-po,pgh-po}
Type T2 = {bos-airport,la-airport,pgh-airport}
Type T3 = {bos,la,pgh}
Type T4 = {package1,package2,package3,package4,package5,package6,
   package7,package8}
Type T5 = {airplane1,airplane2}


TIM: STATE INVARIANTS:

FORALL x:T0 U T4 U T5. FORALL y1. FORALL z1. at(x,y1)
AND at(x,z1) => y1 = z1
FORALL x:T0 U T4 U T5. FORALL y1. FORALL z1. in(x,y1)
AND in(x,z1) => y1 = z1
FORALL x:T0 U T4 U T5. (Exists y1:T1 U T2. at(x,y1)
```

```
OR Exists y1:T0 U T5. in(x,y1))
FORALL x:T0 U T4 U T5. NOT (Exists y1:T1 U T2. at(x,y1)
AND Exists y1:T0 U T5. in(x,y1))
```

```
TIM: DOMAIN INVARIANTS:

|{x0: airplane(x0)}| = 2
|{x0: airport(x0)}| = 3
|{x0: city(x0)}| = 3
|{(x0,x1): in-city(x0,x1)}| = 6
|{x0: location(x0)}| = 6
|{x0: obj(x0)}| = 8
|{x0: truck(x0)}| = 3
```

```
TIM: ATTRIBUTE SPACES:

Objects, x, in T1 U T2 can have property:
Exists y1:T0 U T4 U T5. at(y1,x);
Objects, x, in T0 U T5 can have property:
Exists y1:T0 U T4 U T5. in(y1,x);
Objects, x, in T5 all have property: airplane(x);
Objects, x, in T2 all have property: airport(x);
Objects, x, in T3 all have property: city(x);
Objects, x, in T1 U T2 all have property: Exists y1:T3. in-city(x,y1);
Objects, x, in T3 all have property: Exists y1:T1 U T2. in-city(y1,x);
Objects, x, in T1 U T2 all have property: location(x);
Objects, x, in T4 all have property: obj(x);
Objects, x, in T0 all have property: truck(x);
```

```
TIM: OPERATOR PARAMETER RESTRICTIONS:

drive(x1:T0,x2:T1 U T2,x3:T1 U T2,x4:T3)
fly(x1:T5,x2:T2,x3:T2)
unload(x1:T0 U T4 U T5,x2:T0 U T5,x3:T1 U T2)
load-plane(x1:T4,x2:T5,x3:T1 U T2)
load-truck(x1:T4,x2:T0,x3:T1 U T2)
```

```
TIM: ADDITIONAL STATE INVARIANTS, USING SUB-STATE ANALYSIS:
```

The following invariants add the constraints that trucks and airplanes must always be at a location and never loaded into one another.

```
FORALL x:T0. FORALL y1. FORALL z1. at(x,y1) AND at(x,z1) => y1 = z1
FORALL x:T0. (Exists y1:T1 U T2. at(x,y1))

FORALL x:T5. FORALL y1. FORALL z1. at(x,y1) AND at(x,z1) => y1 = z1
FORALL x:T5. (Exists y1:T1 U T2. at(x,y1))
```

## Appendix D. The Rocket Domain

The Rocket domain used in the construction of Figure 6 is as follows:

```
(define (domain rocket)
        (:predicates    (at ?x ?y)
                        (in ?x ?y)
                        (fuelled ?x)
                        (unfuelled ?x)
                        (loc ?x)
                        (obj ?x)
                        (container ?x))

        (:action fly
                :parameters (?x ?y ?z)
                :precondition (and (at ?x ?y) (loc ?z) (fuelled ?x))
                :effect (and (not (at ?x ?y)) (at ?x ?z) (unfuelled ?x)
                        (not (fuelled ?x))))

        (:action load
                :parameters (?x ?y ?z)
                :precondition (and (obj ?x) (container ?y) (at ?x ?z)
                                                                  (at ?y ?z))
                :effect (and (in ?x ?y) (not (at ?x ?z))))

        (:action unload
                :parameters (?x ?y ?z)
                :precondition (and (at ?y ?z) (in ?x ?y))
                :effect (and (at ?x ?z) (not (in ?x ?y)))))
```

## Appendix E. Operator Test Domain

This domain is an artificial domain used to test the effects of increasing operators and literals in the domain encoding on the performance of TIM. This example is the third instance - the variation was achieved by adding more operator schemas in the pattern of those included here.

```
(define (domain od)
        (:predicates
        (p1 ?x ?y) (q1 ?x ?y)
        (p2 ?x ?y) (q2 ?x ?y)
        (p3 ?x ?y) (q3 ?x ?y)
        (p4 ?x ?y) (q4 ?x ?y)
        (p5 ?x ?y) (q5 ?x ?y)
        (p6 ?x ?y) (q6 ?x ?y)
        (p7 ?x ?y) (q7 ?x ?y)
        (p8 ?x ?y) (q8 ?x ?y)
        (p9 ?x ?y) (q9 ?x ?y)
        (p10 ?x ?y) (q10 ?x ?y)
        (p11 ?x ?y) (q11 ?x ?y)
        (p12 ?x ?y) (q12 ?x ?y)
        (p13 ?x ?y) (q13 ?x ?y)
        (p14 ?x ?y) (q14 ?x ?y)
        (p15 ?x ?y) (q15 ?x ?y)
        (p16 ?x ?y) (q16 ?x ?y)
        (p17 ?x ?y) (q17 ?x ?y)
        (p18 ?x ?y) (q18 ?x ?y)
        (p19 ?x ?y) (q19 ?x ?y)
        (p20 ?x ?y) (q20 ?x ?y))

        (:action o1
                :parameters (?x ?y ?z)
                :precondition (and (p1 ?x ?y) (q1 ?x ?z))
                :effect (and (not (p1 ?x ?y)) (not (q1 ?x ?z))
                                (p1 ?x ?z) (q1 ?x ?y)))

        (:action o2
                :parameters (?x ?y ?z)
                :precondition (and (p2 ?x ?y) (q2 ?x ?z))
                :effect (and (not (p2 ?x ?y)) (not (q2 ?x ?z))
                                (p2 ?x ?z) (q2 ?x ?y)))

        (:action o3
                :parameters (?x ?y ?z)
                :precondition (and (p3 ?x ?y) (q3 ?x ?z))
                :effect (and (not (p3 ?x ?y)) (not (q3 ?x ?z))
                                (p3 ?x ?z) (q3 ?x ?y))))
```

The problem instance was fixed as follows:

```
(define (problem op)
        (:domain od)
        (:objects a b c)
```

```
(:init  (p1 a b)                (q1 a c)
        (p2 a b)                (q2 a c)
        (p3 a b)                (q3 a c)
        (p4 a b)                (q4 a c)
        (p5 a b)                (q5 a c)
        (p6 a b)                (q6 a c)
        (p7 a b)                (q7 a c)
        (p8 a b)                (q8 a c)
        (p9 a b)                (q9 a c)
        (p10 a b)               (q10 a c)
        (p11 a b)               (q11 a c)
        (p12 a b)               (q12 a c)
        (p13 a b)               (q13 a c)
        (p14 a b)               (q14 a c)
        (p15 a b)               (q15 a c)
        (p16 a b)               (q16 a c)
        (p17 a b)               (q17 a c)
        (p18 a b)               (q18 a c)
        (p19 a b)               (q19 a c)
        (p20 a b)               (q20 a c))
(:goal (and (p1 a c) (q1 a b))))
```